\documentclass{article}

\PassOptionsToPackage{numbers,compress}{natbib}
 \usepackage[main, final]{neurips_2026}

\usepackage[utf8]{inputenc} 
\usepackage[T1]{fontenc}    
\usepackage{hyperref}       
\usepackage{url}            
\usepackage{booktabs}       
\usepackage{amsfonts}       
\usepackage{nicefrac}       
\usepackage{microtype}      
\usepackage{xcolor}         
\usepackage{amsmath}
\usepackage{graphicx} 
\usepackage{booktabs}
\usepackage{multirow}
\usepackage{colortbl}
\usepackage{makecell}
\usepackage{float}
\definecolor{oursblue}{RGB}{219,234,254}   
\definecolor{bestgold}{RGB}{254,243,199}   
\definecolor{oomgray}{RGB}{220,220,220}

\title{Rethinking the State Update Gate for Long-Sequence Recurrent 3D Reconstruction}

\author{%
  Kejun Ren$^{1*}$ \quad Lei Jin$^{1*}$ \quad Tianxin Huang$^{2}$ \quad Lianming Xu$^{1}$ \quad Li Wang$^{1\dagger}$ \\[0.4em]
  $^{1}$Beijing University of Posts and Telecommunications, Beijing, China \\
  $^{2}$School of Computing and Data Science, The University of Hong Kong
}

\begin{document}

\maketitle
{\renewcommand{\thefootnote}{}\footnotetext{$^{*}$Equal contribution. \quad $^{\dagger}$Corresponding author.}}

\begin{figure*}[h]
\centering
\includegraphics[width=1.0\linewidth]{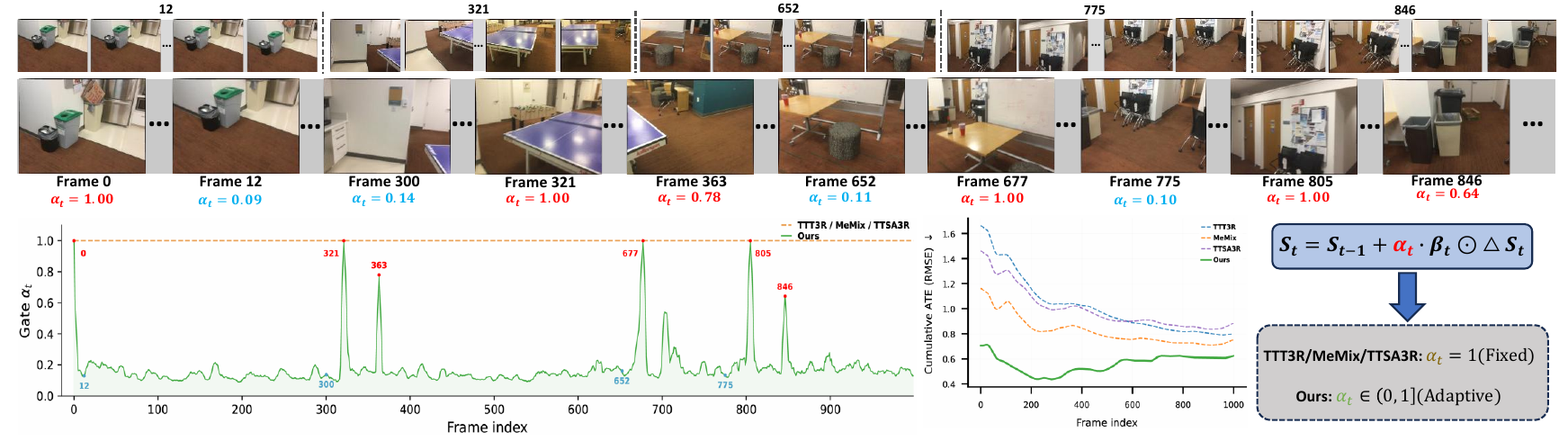}

\caption{\textbf{Adaptive frame gating for streaming 3D reconstruction.}
Existing methods (TTT3R, MeMix, TTSA3R) update the recurrent state $S_t$ with $\alpha_t \equiv 1$ regardless of frame content, accumulating drift on long sequences. We introduce an adaptive frame gate $\alpha_t \in (0,1]$ scaled by frame-to-frame feature change, requiring no parameters, training, or extra forward pass.
The gate shown is AFG-Pose (Sec.~\ref{subsec:two_gates}). \textbf{Top:} Representative frames from a $1000{+}$-frame indoor sequence. \textcolor{red}{Red} keyframes (e.g., \#321, \#805) at significant scene or viewpoint transitions receive large $\alpha_t$ ($\to 1$); \textcolor{cyan!70!black}{cyan} near-duplicates (e.g., \#12, \#300) receive small $\alpha_t$ ($\sim 0.1$) and are strongly down-weighted (but never discarded).
\textbf{Bottom-left:} Gate $\alpha_t$ over the sequence---low during stable motion, spiking at keyframes; existing methods stay at $\alpha_t \equiv 1$.
\textbf{Bottom-middle:} Cumulative ATE---our method maintains a significantly lower and more stable error profile on long sequences, indicating effective drift suppression.
\textbf{Bottom-right:} Our update rule (Eq.~\ref{eq:our_update}): $\alpha_t$ multiplies TTT3R's per-token gate $\beta_t$ at the state-update site.}
\label{fig:teaser}
\end{figure*}

\begin{abstract}
Streaming 3D reconstruction under a strict constant-memory budget hinges on how the recurrent state is updated as the stream evolves. We profile TTT3R-style per-token gates across five benchmarks and discover a structural bottleneck: the gate is intrinsically bounded in magnitude (median $0.31$; never exceeding $0.6$) and nearly frame-invariant, yielding an effective memory horizon of only $\sim$3 frames per state token, which serves as the structural origin of long-sequence drift. We trace this to a missing axis: existing inference-time methods modulate updates only at the per-token, intra-frame level, while the orthogonal frame-level question of \emph{how strongly each frame should contribute to the state} has been treated as content-independent. We close this gap with a scalar frame-level gate $\alpha_t \in (0, 1]$ derived in closed form from frame-to-frame changes of internal features---a graded write weight, inspired by classical Simultaneous Localization and Mapping (SLAM) keyframe selection, that never discards a frame and requires no parameters, no training, and no extra forward pass. Across six benchmarks spanning camera pose, video depth, and 3D reconstruction at sequence lengths up to $4,661$ frames, our gate cuts ATE by $51\%$ on long TUM-RGBD pose sequences, reduces AbsRel by $13.0\%$ on Bonn video depth, and on KITTI long-sequence pose estimation surpasses both LongStream and Keyframe-VO in average ATE, while retaining strictly constant memory at zero training cost. Code is available at \url{https://github.com/rkj12345/AFG}.
\end{abstract}

\section{Introduction}
\label{intro}

Streaming 3D reconstruction processes an RGB video incrementally and produces camera poses and pixel-aligned geometry on the fly, under strict memory and latency constraints that grow with the input length~\citep{wang2025continuous,zhuo2025streaming}. Offline batch methods such as DUSt3R~\cite{wang2024dust3r}, MASt3R~\cite{leroy2024grounding}, VGGT~\cite{wang2025vggt}, and Fast3R~\cite{yang2025fast3r} produce strong reconstructions through global attention over all frames, but their quadratic cost and the requirement that all frames be loaded simultaneously preclude online use on long sequences. Among streaming approaches, CUT3R~\cite{wang2025continuous} achieves a strictly constant memory footprint by compressing the full sequence history into a fixed-size recurrent state of $N$ tokens, $S \in \mathbb{R}^{N \times d}$, that each incoming frame reads from and writes back into. KV-cache variants~\cite{zhuo2025streaming,wu2025point3r} retain richer history but grow with sequence length, while sliding-window attention~\cite{li2025wint3r} keeps memory constant only by discarding early frames, making CUT3R the unique combination of constant memory and full-history retention. \textbf{Under a strict constant-memory budget, the central question becomes how this state should be updated as the stream evolves} (Fig.~\ref{fig:teaser}).

Recent inference-time methods all modulate this update along the same axis: a per-token gate $\beta_t \in [0,1]^N$ that decides which state slots absorb the current frame's residual. TTT3R~\cite{chen2025ttt3r} derives $\beta_t$ from cross-attention logits as a constrained test-time gradient step; TTSA3R~\cite{zheng2026ttsa3r} reshapes it via attention-style gating; the concurrent MeMix~\cite{dong2026memix} adds a hard token mask. Through systematic profiling on long sequences across five benchmarks, we discover a striking regularity: $\beta_t$ is structurally bounded (median $0.31$; no value exceeds $0.6$), nearly invariant across datasets, and---most importantly---its variation \emph{across} frames is less than half of its variation \emph{within} a frame. Under an EMA reading of the update, this corresponds to an effective memory horizon of only $\sim$3 frames, structurally explaining the long-sequence drift these methods continue to exhibit. \textbf{Yet a more fundamental observation lies beneath this bound: all of these designs lack any frame-level adaptivity. They modulate updates only along the intra-frame axis, i.e., per-token within a single frame, and treat every frame as equally informative regardless of its content.}

Classical SLAM has, for two decades, treated content-dependent admission as a first-class problem under the name \emph{keyframe selection}~\cite{campos2021orb}: only frames bringing sufficient novelty enter the map, while redundant ones are skipped. Modern streaming reconstruction has quietly abandoned this idea, every $I_t$ writes into the state with the same nominal strength regardless of content. We argue that this is the missing axis, and propose to reintroduce it not as a binary admit/skip decision but as a \emph{graded write weight}: every frame still contributes, but with a strength that grows with its novelty. We borrow the signal of keyframe selection (how new a frame looks) rather than its role: keyframe selection bounds computation by choosing which frames enter a growing map, whereas our state has a fixed size and the problem is how strongly each frame overwrites it (Sec.~\ref{subsec:kf_relation}). Concretely, we introduce a scalar frame-level gate $\alpha_t \in (0, 1]$ that multiplicatively modulates the state-update residual. Computed in closed form from the frame-to-frame change of internal features the base model already produces, $\alpha_t$ adds no parameters and no extra forward pass. Most frames are redundant and receive small $\alpha_t$ that barely perturbs the state; only significant scene or viewpoint transitions drive $\alpha_t$ toward $1$, where the update proceeds at full TTT3R strength.

The design of Adaptive Frame Gating (AFG) is minimal: a single multiplicative scalar at the state-update site. As a plug-in modifier, AFG preserves the constant-memory inference property of CUT3R-style backbones, unlike retraining-based alternatives such as LongStream~\cite{cheng2026longstream}. Across six benchmarks (TUM-RGBD, ScanNet, Bonn, KITTI, 7-Scenes, and NRGBD) with sequences from $50$ to $4661$ frames, AFG improves over inference-time gating baselines including TTT3R, MeMix, and TTSA3R on long sequences; on long TUM-RGBD trajectories ($L \geq 600$) it reduces ATE by $51\%$ over TTT3R, and on KITTI it surpasses both LongStream and the RL-trained Keyframe-VO~\cite{dai2026keyframe}.

Our contributions are summarized as follows.
\begin{itemize}
    \item \textbf{Analysis of per-token gating.}
    We identify a key limitation of TTT3R-style per-token gates: they are nearly frame-invariant and imply an effective memory horizon of only $\sim$3 frames, suggesting that the lack of frame-level adaptivity is a key bottleneck in long-sequence streaming reconstruction.

    \item \textbf{Adaptive Frame Gating.}
    We propose AFG, a scalar frame-level gate $\alpha_t \in (0,1]$ computed in closed form from features already produced by the base model. We instantiate it as AFG-Img and AFG-Pose, using encoder global features and decoder pose tokens, respectively. Both variants are parameter-free, training-free, require no extra forward pass, and preserve constant memory.

    \item \textbf{Extensive long-sequence evaluation.}
    We validate AFG across pose, depth, and reconstruction tasks on six benchmarks. On long sequences it improves over inference-time gating baselines and compares favorably with long-sequence methods based on retraining or learned keyframe policies.
\end{itemize}

\section{Related Work}

\paragraph{Feed-forward 3D reconstruction.}
Feed-forward methods bypass the conventional Structure-from-Motion (SfM) pipeline by directly regressing scene geometry from input images. DUSt3R~\cite{wang2024dust3r} and MASt3R~\cite{leroy2024grounding} cast pairwise reconstruction as dense pointmap regression in a shared coordinate frame; VGGT~\cite{wang2025vggt} and Fast3R~\cite{yang2025fast3r} extend this to the multi-view setting through global Transformer attention. While these methods achieve high reconstruction quality, they require all input frames to reside in attention simultaneously, scaling quadratically with sequence length and precluding online use on long videos.

\paragraph{Streaming 3D reconstruction.}
To process inputs of arbitrary length, streaming methods turn to sequential or recurrent paradigms. Spann3R~\cite{wang20253d} introduces an explicit spatial memory for incremental reconstruction; CUT3R~\cite{wang2025continuous} formulates the problem as an RNN with a fixed-size recurrent state---the only design among streaming variants that strictly preserves constant memory. Subsequent works explore richer memory structures, including KV caches~\cite{zhuo2025streaming}, point-level memory~\cite{wu2025point3r}, and sliding-window attention~\cite{li2025wint3r}, but their memory grows with sequence length and limits deployment on long videos. Within the constant-memory regime, how the recurrent state should be updated as the stream evolves remains an open question, since unfiltered residual accumulation has been shown to cause severe long-sequence drift~\cite{chen2025ttt3r}.
\paragraph{Long-sequence adaptation strategies.}

Three complementary lines of work address this long-sequence degradation. The first relies on \emph{architecture-level retraining}: a growing body of work retrains new architectures or memory mechanisms tailored for long-sequence streaming reconstruction~\cite{cheng2026longstream,chen2025long3r,zhang2026loger,jin2026zipmap,xie2026scal3r,liu2026mem3r,xu2026pas3r}. As a representative, LongStream~\cite{cheng2026longstream}, built on the VGGT family, redesigns camera pose parameterization and KV-cache training to mitigate first-frame anchor bias and cache saturation. These approaches share the cost of giving up the constant-memory property and training from scratch. The second uses \emph{external frame selection policies}: Keyframe-VO~\cite{dai2026keyframe} freezes the base model and trains a reinforcement-learning policy that makes a discrete admit/skip decision for each frame, extending classical SLAM keyframe selection~\cite{mur2015orb,campos2021orb} to feed-forward 3D models. The third modulates state updates at \emph{inference time}: TTT3R~\cite{chen2025ttt3r} reinterprets the recurrent update as a constrained test-time gradient step~\cite{sun2023learning} and derives per-token learning rates $\beta_t \in [0,1]^N$ from cross-attention logits; TTSA3R~\cite{zheng2026ttsa3r} reshapes $\beta_t$ via attention-style gating; the concurrent MeMix~\cite{dong2026memix} adds a per-token hard mask. Inference-time methods preserve constant memory and require no training, but all operate exclusively on the per-token, intra-frame axis. \textbf{Our work belongs to this third line and adds the orthogonal frame-level axis}: compared to TTT3R, TTSA3R, and MeMix, we go beyond per-token modulation by making each frame's contribution to the state content-dependent at the frame level; compared to Keyframe-VO, we share the goal of content-dependent admission but replace its discrete RL-trained admit/skip policy with a graded write weight computed in closed form from the base model's internal features---requiring no training, no parameters, and no extra forward pass.

\section{Structural Diagnosis of Per-Token Gating}
\label{sec:analysis}

\subsection{Recurrent State Updates for Continuous 3D Reconstruction}
\label{subsec:bg}

Given a continuous image stream $\{I_t\}_{t=1}^{T}$, online 3D reconstruction estimates per-frame camera pose $T_t$, intrinsics $K_t$, and pixel-aligned pointmap $P_t$. CUT3R~\cite{wang2025continuous} casts this as a recurrent sequence model with a fixed-length state $S \in \mathbb{R}^{N \times d}$ initialized from learnable embeddings: each frame $I_t$ is tokenized into image tokens $X_t$, the state is updated as $S_t = \mathrm{Update}(S_{t-1}, X_t)$, and a read-out $Y_t = \mathrm{Read}(S_t, X_t)$ is decoded into $(T_t, K_t, P_t)$ by a prediction head.

\paragraph{State--input interaction.}
$\mathrm{Update}$ and $\mathrm{Read}$ are jointly realized by an $L$-layer dual-stream cross-attention decoder. The state stream attends to image tokens at each layer:
\begin{equation}
S^{(l)} = S^{(l-1)} + \mathrm{softmax}\!\left( Q_S^{(l)} {K_X^{(l)}}^{\!\top} \right) V_X^{(l)},
\label{eq:dec_layer}
\end{equation}
with a symmetric stream updating $X^{(l)}$ from $S^{(l-1)}$. Stacking $L$ layers yields the per-frame update $S_t = S_{t-1} + \Delta S_t$, where $\Delta S_t$ aggregates the cross-attention contributions of the current frame.

\paragraph{Continuous state update (CUT3R).}
CUT3R writes this residual back unconditionally: $S_t = S_{t-1} + \Delta S_t$. On long sequences with redundant content, this monotonic accumulation overwrites earlier observations and drives the state toward catastrophic drift~\cite{chen2025ttt3r}.

\paragraph{Test-time learning for state update (TTT3R).}
TTT3R~\cite{chen2025ttt3r} reinterprets $\Delta S_t$ as a gradient step in a test-time training framework~\cite{sun2023learning}, with $S$ acting as model parameters, $X_t$ as a test sample, and $\Delta S_t$ as a descent direction on a self-supervised reconstruction loss. To regulate the step per state token, TTT3R averages the cross-attention logits over $L$ decoder layers, $H$ heads per layer, and $K$ image tokens per frame, and applies the resulting per-token learning rate multiplicatively before integration:
\begin{equation}
\beta_t \;=\; \sigma\!\left( \frac{1}{LHK} \sum_{l, h, k} Q_{S_{t-1}}^{(l, h)} {K_{X_t}^{(l, h, k)}}^{\!\top} \right) \;\in\; [0, 1]^N,
\qquad
S_t = S_{t-1} + \beta_t \odot \Delta S_t.
\label{eq:tt3r_beta}
\end{equation}
We write $\beta_t^{(n)}$ for the gate value at frame $t \in \{1,\dots,T\}$ on the $n$-th state token, $n \in \{1,\dots,N\}$. Each state token now retains history in proportion to its current cross-attention relevance, mitigating the drift of CUT3R. Whether $\beta_t$ provides effective control on long sequences---an empirical question not examined in prior work---is the focus of this section.

\subsection{Empirical Properties of $\beta$: Bounded Magnitude and Negligible Frame Variation}
\label{subsec:fact1}

\begin{figure}[t]
\centering
\includegraphics[width=\linewidth]{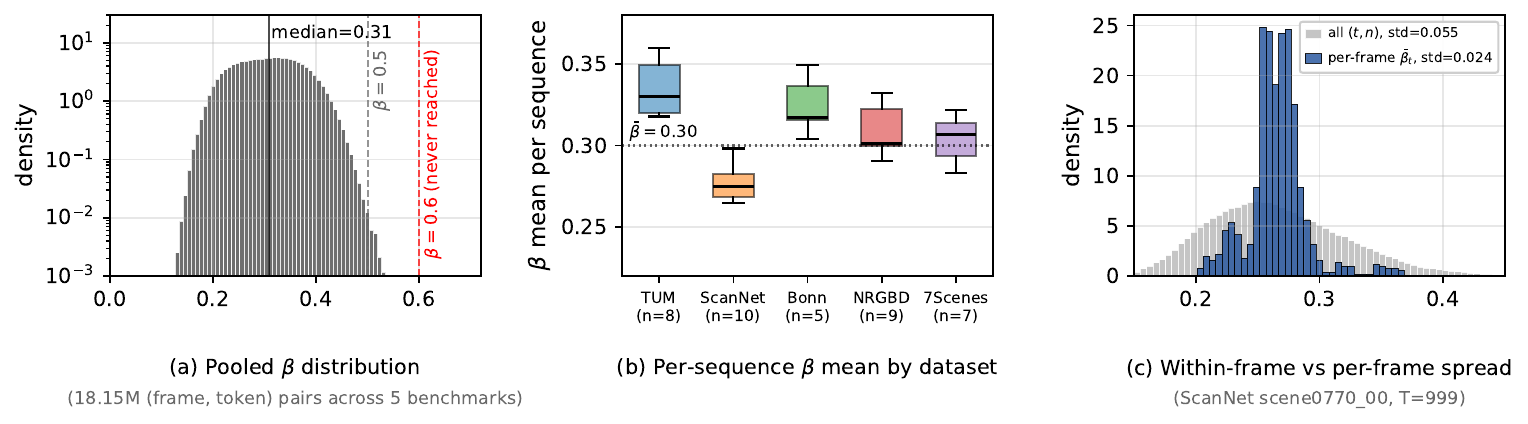}
\caption{\textbf{Empirical properties of TTT3R's per-token learning rate $\beta$.}
\textbf{(a)} Density of $\beta$ pooled over all $(t, n)$ pairs from 39 sequences across 5 benchmarks (18.15M points). The distribution is sharply concentrated; \emph{no} measurement exceeds $\beta=0.6$.
\textbf{(b)} Per-sequence mean of $\beta$ grouped by dataset; despite the diverse motion regimes, all five benchmarks lie within $[0.27, 0.34]$.
\textbf{(c)} For a representative long sequence (ScanNet \texttt{scene0770\_00}, $T=999$), the histogram of all $(t, n)$ values (gray, $\mathrm{std}=0.055$) is more than twice as wide as the histogram of per-frame means $\bar\beta_t$ (blue, $\mathrm{std}=0.024$): TTT3R exhibits per-token variation, but the per-frame mean is essentially constant.}
\label{fig:beta_dist}
\end{figure}

We profile $\beta_t$ across 5 benchmarks (TUM-RGBD, ScanNet, Bonn, NRGBD, 7-Scenes) covering indoor scanning, handheld, and dynamic scenes. Across all sequences and frames, we collect $\beta_t^{(n)}$ at every $(t, n)$ pair and analyze its distribution along two axes: pooled magnitude (Fig.~\ref{fig:beta_dist}a, b) and within-frame versus per-frame variation (Fig.~\ref{fig:beta_dist}c).

\paragraph{$\beta$ is structurally bounded.}
On the pooled distribution (Fig.~\ref{fig:beta_dist}a), $\beta$ has median $0.31$, $99\%$ of its mass below $0.44$, and a maximum of $0.558$; \emph{none} of the $18.15$M measurements exceeds $0.6$. This bound holds uniformly across datasets (Fig.~\ref{fig:beta_dist}b): per-sequence means lie in a tight band of $[0.27, 0.34]$, with no benchmark significantly higher or lower than the global mean of $0.31$. The bound is therefore a property of the architecture rather than of any particular dataset.

\paragraph{$\beta$ has negligible frame-to-frame variation.}
Within a single sequence, the spread of $\beta$ across all $(t, n)$ pairs is more than twice as large as the spread of the per-frame means $\bar\beta_t = \frac{1}{N}\sum_n \beta_t^{(n)}$ (Fig.~\ref{fig:beta_dist}c): on ScanNet \texttt{scene0770\_00}, the within-frame std is $0.055$ while the cross-frame std of $\bar\beta_t$ is only $0.024$. The same pattern holds across the other benchmarks: the per-frame mean of $\beta$ remains essentially constant regardless of sequence length or motion regime. While the per-token gate yields meaningful variation across the $N$ state tokens (consistent with TTT3R's design intent), the per-frame mean is essentially constant: TTT3R provides token-level selectivity but \emph{no} frame-level adaptivity.

\paragraph{Why $\beta$ behaves this way.}
Both properties stem from the multi-axis averaging in Eq.~\ref{eq:tt3r_beta}: each $\beta_t^{(n)}$ is the sigmoid of an average over $L \times H \times K$ ($\approx 110$K in the CUT3R-512 architecture) cross-attention logits. This averaging has two effects: (i) extreme logit values are diluted into a narrow effective range, and (ii) the summation over $K$ image tokens fuses frame-specific spatial information into a global statistic, with the residual mean dominated by the pretrained $(Q, K)$ projections rather than by $X_t$. Let $\bar A$ denote the averaged cross-attention logit inside the sigmoid of Eq.~\ref{eq:tt3r_beta}, i.e., $\bar A = \tfrac{1}{LHK}\sum_{l,h,k} Q_{S_{t-1}}^{(l,h)} {K_{X_t}^{(l,h,k)}}^{\!\top}$. Empirically, $\bar A \approx -0.83$ across all measurements, yielding $\sigma(\bar A) \approx 0.30$ and matching the three observations of Fig.~\ref{fig:beta_dist}. The consequence is that every frame contributes a similar moderate fraction ($\sim$30\%) of its residual $\Delta S_t$ to the state regardless of content. We next examine what this implies for memory over long sequences.

\subsection{Memory Horizon Implied by Bounded $\beta$}
\label{subsec:fact3}

\paragraph{EMA decay and memory horizon.}
Writing the residual as $\Delta S_t = \tilde S_t - S_{t-1}$, where $\tilde S_t$ is the state the current frame would write on its own, the update of Eq.~\ref{eq:tt3r_beta} is a convex combination, $S_t = (1-\beta_t)\odot S_{t-1} + \beta_t \odot \tilde S_t$, i.e., exactly an exponential moving average on each state slot with smoothing factor $\beta_t^{(n)}$. The only assumption we add is that $\beta$ is approximately constant in time, which is what Sec.~\ref{subsec:fact1} measures. The contribution of frame $t-k$ to $S_t^{(n)}$ then decays geometrically as $(1-\beta)^k$, giving an effective memory horizon of approximately $1/\beta$ frames\footnote{Defined as the lag at which the contribution drops to $1/e$ of its instantaneous value: $(1-\beta)^k = 1/e \Rightarrow k \approx 1/\beta$ for small $\beta$.}. Since the per-frame mean is essentially constant across datasets ($\bar\beta_t \in [0.27, 0.34]$, Sec.~\ref{subsec:fact1}):
\begin{equation}
\mathrm{horizon} \;\approx\; 1/\bar\beta \;\approx\; 3\ \text{frames}.
\label{eq:horizon}
\end{equation}
This horizon is not only short but \emph{stable}: it stays within $[2.9, 3.7]$ regardless of sequence length or motion regime. Over a sequence of $T=1000$ frames, the contribution of frame $0$ to $S_{100}$ has already decayed to $(0.69)^{100} \approx 10^{-16}$---early observations are erased almost immediately and never recovered.

The horizon is an exact property of the update rule; the step from a short horizon to pose drift is not derived but established empirically, by the benchmarks of Sec.~\ref{sec:experiments} and the controlled probe of Appendix~\ref{app:redundancy}. We argue that this short horizon is the structural cause of long-sequence drift: since $\beta$ is content-independent (Sec.~\ref{subsec:fact1}), every frame overwrites the state with equal strength, so an informative keyframe is displaced within $\sim$3 steps by whatever follows---including near-redundant frames. The memory is not just short, but \emph{indiscriminately} short: early geometric evidence is forgotten long before it ceases to be relevant. We verify this mechanism via a controlled redundancy experiment in Appendix~\ref{app:redundancy}.

\section{Adaptive Frame Gating}
\label{sec:method}

Building on the diagnosis of Sec.~\ref{sec:analysis}, we restore the missing frame-level adaptivity with a scalar gate $\alpha_t$ that adaptively scales $\beta_t$ based on observation novelty.

\subsection{Factorized State Update}
\label{subsec:method_decomp}

We replace the TTT3R update of Eq.~\ref{eq:tt3r_beta} with the factorized form
\begin{equation}
S_t \;=\; S_{t-1} \;+\; \alpha_t \cdot \beta_t \odot \Delta S_t,
\label{eq:our_update}
\end{equation}
where $\alpha_t \in (0, 1]$ is a scalar frame-level gate that adaptively scales the current frame's contribution to the state based on its content, and $\beta_t \in [0, 1]^N$ is TTT3R's per-token gate from Eq.~\ref{eq:tt3r_beta} that determines \emph{which} state tokens should receive the contribution. The two gates operate on orthogonal axes, token (spatial) and frame (temporal), and combine multiplicatively: $\beta_t^{(n)}$ controls per-token spatial selection while $\alpha_t$ modulates the per-frame update strength. This composition preserves TTT3R's token selectivity while adding an independent temporal modulation. 

\subsection{Frame Gate from CUT3R-Internal Features}
\label{subsec:two_gates}

We compute $\alpha_t$ from the frame-to-frame change of a feature that CUT3R already produces, yielding two natural instances. Both add zero parameters and zero forward passes; throughout, $\sigma(\cdot)$ denotes the logistic sigmoid $\sigma(z) = 1/(1+e^{-z})$.

\paragraph{AFG-Img.} We use the encoder's per-frame global feature $g_t \in \mathbb{R}^{d}$, defined as the spatial mean of the encoder's patch tokens, $g_t = \tfrac{1}{K}\sum_{k=1}^{K} X_t^{(k)}$, which CUT3R already employs as its LocalMemory query. Its frame-to-frame change drives:
\begin{equation}
\alpha^{\text{img}}_t \;=\; \sigma\!\left( \kappa_{\text{img}}\big(\|g_t - g_{t-1}\|_2 - \tau_{\text{img}}\big) \right).
\label{eq:img_gate}
\end{equation}

\paragraph{AFG-Pose.} We use the first row of the final decoder-layer state, $p_t = S^{(L)}_{t,1} \in \mathbb{R}^{d}$, which feeds CUT3R's pose prediction head and captures the model's per-frame pose representation. Its frame-to-frame change drives:
\begin{equation}
\alpha^{\text{pose}}_t \;=\; \sigma\!\left( \kappa_{\text{pose}}\big(\|p_t - p_{t-1}\|_2 - \tau_{\text{pose}}\big) \right).
\label{eq:pose_gate}
\end{equation}

Here $\tau$ is the centre of the sigmoid (a threshold on the feature change) and $\kappa$ its sharpness; the two constants differ between variants because $g_t$ and $p_t$ live on different scales. We use $(\kappa_{\text{img}}, \tau_{\text{img}}) = (1, 1.5)$ and $(\kappa_{\text{pose}}, \tau_{\text{pose}}) = (3, 1.0)$ on every benchmark, with no per-scene tuning; the sensitivity to both constants is reported in Appendix~\ref{app:gate_constants}. When the feature change is zero (a frame identical to its predecessor), $\alpha_t$ saturates at the strict positive lower bound $\sigma(-\kappa\tau) > 0$---$\sigma(-1.5)\approx0.18$ for AFG-Img and $\sigma(-3)\approx0.047$ for AFG-Pose---rather than collapsing to zero, so the gate remains in $(0, 1]$ on arbitrarily long redundant runs. The gate is capped at $1$ by construction, so AFG can only weaken a write, never enlarge it beyond TTT3R's; we test this design choice in Appendix~\ref{app:cap}. The two variants capture novelty at different abstraction levels: $g_t$ reflects raw visual content, $p_t$ reflects the model's pose-related cognition. Empirically, AFG-Img excels on smooth-scanning regimes where pose changes are gentle, while AFG-Pose excels on dynamic regimes where pose changes carry the strongest signal (Sec.~\ref{sec:experiments}). We use them as complementary regime-specific variants, since naive fusion does not exceed the better single gate (Sec.~\ref{subsec:exp_ablation}).

\paragraph{Interpretation: a longer, content-aware horizon.}
$\alpha_t$ acts as an \emph{inference-time learning rate} on the recurrent state: $\alpha_t \to 1$ on novel frames (absorbed at full TTT3R strength), $\alpha_t \to \alpha_{\min} = \sigma(-\kappa\tau)$ on redundant ones. This reshapes the memory horizon of Sec.~\ref{sec:analysis}: a keyframe is overwritten not at the fixed rate $\beta$, but at $\alpha_t\beta$ set by the novelty of \emph{subsequent} frames. When followed by low-information frames ($\alpha_t \to \alpha_{\min}$), its contribution decays as $(1-\alpha_{\min}\beta)^k$, extending the horizon from $1/\beta\approx3$ frames to $1/(\alpha_{\min}\beta)$. Writing $H(\alpha) \approx 1/(\alpha\bar\beta)$, TTT3R ($\alpha \equiv 1$) has $H \approx 3$ frames whatever the content; with the observed $\alpha_{\min}\approx0.048$ on fully redundant inputs, $H$ reaches $\sim$60 frames---up to a $\sim$20$\times$ extension under sustained redundancy. AFG thus does not remember harder; it \emph{forgets selectively}, spending a limited memory budget on informative frames rather than the most recent ones. We verify this with a controlled redundancy-injection probe in Appendix~\ref{app:redundancy}, where the drift decay measured independently gives $\sim$64 frames.

\subsection{Relation to Keyframe Selection}
\label{subsec:kf_relation}
AFG shares its signal with classical keyframe selection but differs in role and mechanism. Keyframe selection exists because a map and its optimisation grow with the number of frames, so a subset must be chosen and a rejected frame never enters the map. In a fixed-size recurrent state nothing needs to be excluded; the problem is that new frames keep overwriting old ones, so AFG sets \emph{how strongly} each frame writes, and every frame writes. Two consequences follow. First, since $\alpha_t \geq \alpha_{\min} > 0$, a frame whose novelty is under-read is delayed rather than lost, which is why an adjacent-frame change suffices as the signal, whereas a binary rule must measure novelty against the map to avoid irrecoverable mistakes. Second, the difference is measurable (Appendix~\ref{app:keyframe}): binarising our gate with non-keyframes set to $0$, i.e., the classical admit/skip rule, raises TUM-RGBD ATE from $0.052$ to $0.061$ and RPE$_r$ from $0.647$ to $0.880$ relative to keeping them at a small weight, and a hard response is worse than our graded one on ScanNet ($0.352$ vs.\ $0.281$ ATE for AFG-Pose). Measuring novelty against the last accepted frame instead of the previous frame, closer to classical keyframe criteria, is also worse ($0.065$ vs.\ $0.054$ ATE; $0.092$ vs.\ $0.086$ AbsRel).

\section{Experiments}
\label{sec:experiments}

\paragraph{Tasks and benchmarks.}
We evaluate on three streaming 3D reconstruction tasks: camera pose estimation on TUM-RGBD~\cite{sturm2012benchmark}, ScanNet~\cite{dai2017scannet}, and KITTI~\cite{geiger2013vision}; video depth estimation on Bonn~\cite{palazzolo2019refusion}; and 3D reconstruction on 7-Scenes~\cite{shotton2013scene} and NRGBD~\cite{azinovic2022neural}. Sequence lengths span $50$ to $4661$ frames. We adopt standard alignment conventions throughout: Sim(3) Umeyama alignment for camera pose, and metric alignment for depth (scale-and-shift depth results are additionally reported in Appendix~\ref{app:bonn_ss} as a robustness check). Per-task evaluation details are given in each subsection.

\paragraph{Baselines.}
We compare against three families: inference-time gating methods (TTT3R~\cite{chen2025ttt3r}, TTSA3R~\cite{zheng2026ttsa3r}, MeMix~\cite{dong2026memix}); feedforward backbones (CUT3R~\cite{wang2025continuous}, StreamVGGT~\cite{zhuo2025streaming}, FastVGGT~\cite{shen2025fastvggt}, InfiniteVGGT~\cite{yuan2026infinitevggt}); and long-sequence-specialized methods (LongStream~\cite{cheng2026longstream}, Keyframe-VO~\cite{dai2026keyframe}).

\paragraph{Implementation details.}
AFG is applied on top of TTT3R to the same frozen CUT3R checkpoint without fine-tuning, with the gate constants of Sec.~\ref{subsec:two_gates} used unchanged across all benchmarks. All experiments are conducted on a single NVIDIA A100 (80\,GB).

\subsection{Camera Pose Estimation}
\label{subsec:exp_pose}

\begin{figure}[H]
  \centering
  \includegraphics[width=0.6\linewidth]{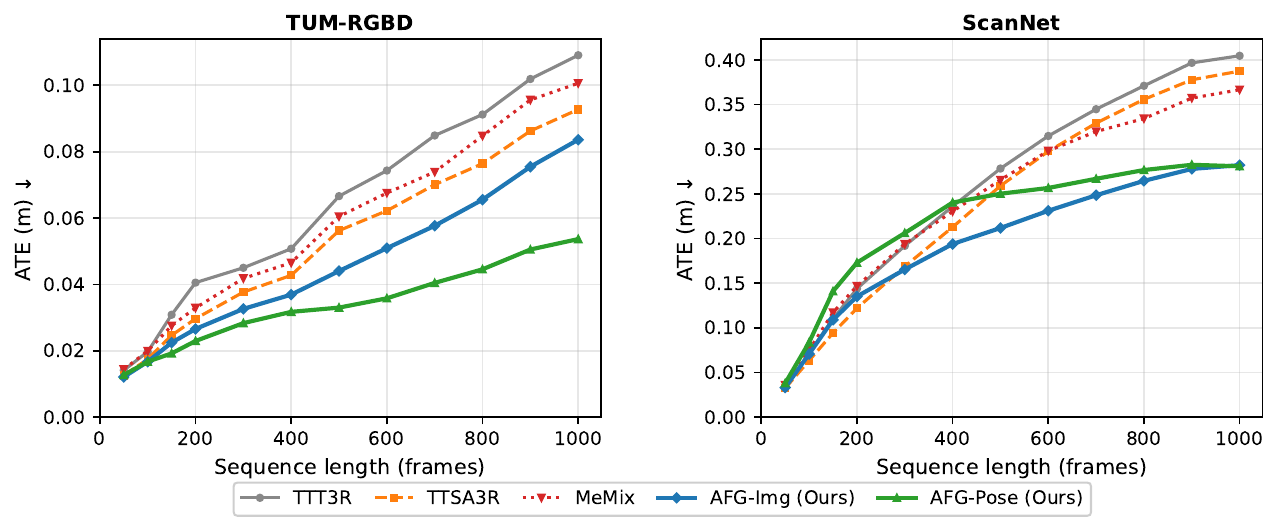}
  \caption{\textbf{Camera pose ATE vs.\ sequence length.}
  On long sequences AFG outperforms TTT3R, TTSA3R, and MeMix on TUM-RGBD and ScanNet, with the gap widening as length grows; on short ScanNet sequences TTSA3R is best.}
  \label{fig:pose_ate}
\end{figure}

Following prior work~\cite{wang2025continuous,chen2025ttt3r,dong2026memix}, we evaluate long-sequence absolute trajectory error (ATE) on TUM-RGBD and ScanNet as sequence length increases; results are shown in Fig.~\ref{fig:pose_ate}, with per-length numerical values tabulated in Appendix~\ref{app:pose_detailed}. On TUM-RGBD both variants outperform TTT3R, TTSA3R, and MeMix for $L \geq 200$; on ScanNet AFG-Img is best for $L \geq 300$, while AFG-Pose trails TTT3R up to $L=400$ and overtakes all baselines from $L=500$. The gap widens as sequences grow longer, and on short ScanNet sequences ($L \leq 200$), where little drift has accumulated, TTSA3R is best; MeMix offers limited additional gain over TTT3R throughout, supporting the analysis of Sec.~\ref{sec:analysis} that token-level gating alone cannot address frame-level drift. AFG-Pose is the stronger variant on TUM-RGBD (dynamic walking sequences), reducing average ATE at $L \geq 600$ by $51\%$ over TTT3R, while AFG-Img leads on ScanNet (smooth indoor scanning) with a $29\%$ reduction.

\subsection{Long-Sequence Pose Estimation on KITTI}
\label{subsec:exp_kitti}

On KITTI Odometry~\cite{geiger2013vision} (Table~\ref{tab:kitti}), we additionally compare against two long-sequence-specialized baselines: the retrained-from-scratch LongStream~\cite{cheng2026longstream} and the RL-policy-trained Keyframe-VO~\cite{dai2026keyframe}. Concurrent inference-time methods TTSA3R~\cite{zheng2026ttsa3r} and MeMix~\cite{dong2026memix} do not report KITTI results and are instead compared on the indoor benchmarks of Sec.~\ref{subsec:exp_pose}. AFG-Pose achieves the lowest average ATE ($43.96$~m), reducing TTT3R by $35.9\%$ and surpassing both long-sequence-specialized baselines---LongStream ($51.90$) and Keyframe-VO ($87.00$)---despite requiring no retraining, no architectural change, and no extra forward pass; AFG-Img also improves TTT3R to $56.60$. Consistent with the EMA horizon analysis in Sec.~\ref{sec:analysis}, gains tend to grow with length rather than with loop structure: on the three sequences longer than $4000$ frames (\texttt{00}/\texttt{02}/\texttt{08}) AFG-Pose reduces ATE over TTT3R by $39$--$51\%$ and on \texttt{09} by $64\%$, while the shortest (\texttt{04}, $L=271$) gets worse. The trend is not strict (\texttt{05}: $+11\%$; \texttt{07}: $+45\%$), and loops do not separate the outcomes: the largest gain (\texttt{09}) and the largest loss (\texttt{07}) are both loop sequences, and among the three $1101$-frame sequences the loop-free \texttt{01} ($-17\%$) sits between the two loop sequences \texttt{06} ($-20\%$) and \texttt{07} ($+45\%$). The per-sequence breakdown is in Appendix~\ref{app:kitti_length}. AFG-Pose beats AFG-Img on $7$ of $11$ sequences, matching the regime mapping of Sec.~\ref{sec:method}; trajectories for all $11$ sequences are shown in Appendix~\ref{app:kitti_traj}.

\begin{table*}[h]
\centering
\caption{\textbf{Long-sequence pose estimation on KITTI~\cite{geiger2013vision}.} ATE $\downarrow$ (m) of the aligned trajectory. Baseline results are taken from the original papers. Best in \textbf{bold}, second-best \underline{underlined}.}
\label{tab:kitti}
\setlength{\tabcolsep}{1.5pt}
\renewcommand{\arraystretch}{0.95}
\scriptsize
\resizebox{\textwidth}{!}{%
\begin{tabular}{l|ccccccccccc|c}
\toprule
\multirow{3}{*}{\textbf{Methods}} & \multicolumn{12}{c}{\textbf{KITTI}~(ATE $\downarrow$)} \\
\cmidrule(lr){2-13}
 & 00 & 01 & 02 & 03 & 04 & 05 & 06 & 07 & 08 & 09 & 10 & \multirow{2}{*}{Avg.} \\
 & \tiny{4541, 3.7km} & \tiny{1101, 2.5km} & \tiny{4661, 5.1km} & \tiny{801, 0.6km} & \tiny{271, 0.4km} & \tiny{2761, 2.2km} & \tiny{1101, 1.2km} & \tiny{1101, 0.7km} & \tiny{4071, 3.2km} & \tiny{1591, 1.7km} & \tiny{1201, 0.9km} & \\
\midrule
CUT3R                     & 190.38 &  90.59 & 264.39 &  20.40 &  7.31 &  92.25 &  67.54 &  22.48 & 145.08 &  67.42 &  40.00 &  91.62 \\
TTT3R                     & 118.20 & 102.98 & 242.05 &  14.38 &  4.51 & \textbf{31.62} &  37.77 & \textbf{12.00} &  81.52 &  79.10 &  29.82 &  68.54 \\
InfiniteVGGT              & 186.46 & 623.62 & 289.16 & 166.74 & 68.00 & 143.84 & 117.57 &  85.33 & 221.36 & 215.41 & 156.92 & 206.78 \\
FastVGGT                  & 102.50 & 176.80 & 170.00 & \underline{11.20} & \underline{2.50} &  76.10 & 131.20 &  61.50 &  99.30 &  99.30 &  36.70 &  87.90 \\
StreamVGGT                & 191.93 & 653.06 & 303.35 & 157.50 & 108.24 & 160.46 & 133.71 &  89.00 & 263.95 & 216.69 & 209.80 & 226.15 \\
\midrule
LongStream                &  92.55 & \textbf{46.01} & \underline{134.70} & \textbf{3.81} & \textbf{1.95} &  84.69 & \textbf{23.12} &  14.93 &  62.07 &  85.61 & \textbf{21.48} & \underline{51.90} \\
Keyframe-VO               & 138.10 & 179.10 & 153.50 &  12.00 & 13.20 & 131.60 &  73.00 &  45.90 &  86.50 &  97.50 & \underline{26.80} &  87.00 \\
\midrule
AFG-Img \textbf{(Ours)}   & \underline{89.03} & \underline{79.40} & 229.25 &  14.68 &  8.47 & \underline{32.21} &  31.29 & \underline{14.01} & \underline{49.61} & \underline{47.21} &  27.44 &  56.60 \\
AFG-Pose \textbf{(Ours)}  & \textbf{72.22} &  85.49 & \textbf{118.98} &  12.07 &  6.49 &  35.16 & \underline{30.15} &  17.44 & \textbf{45.92} & \textbf{28.56} &  31.09 & \textbf{43.96} \\
\bottomrule
\end{tabular}%
}
\end{table*}

\subsection{Video Depth Estimation}
\label{subsec:exp_depth}

\begin{figure}[h]
  \centering
  \includegraphics[width=0.80\linewidth]{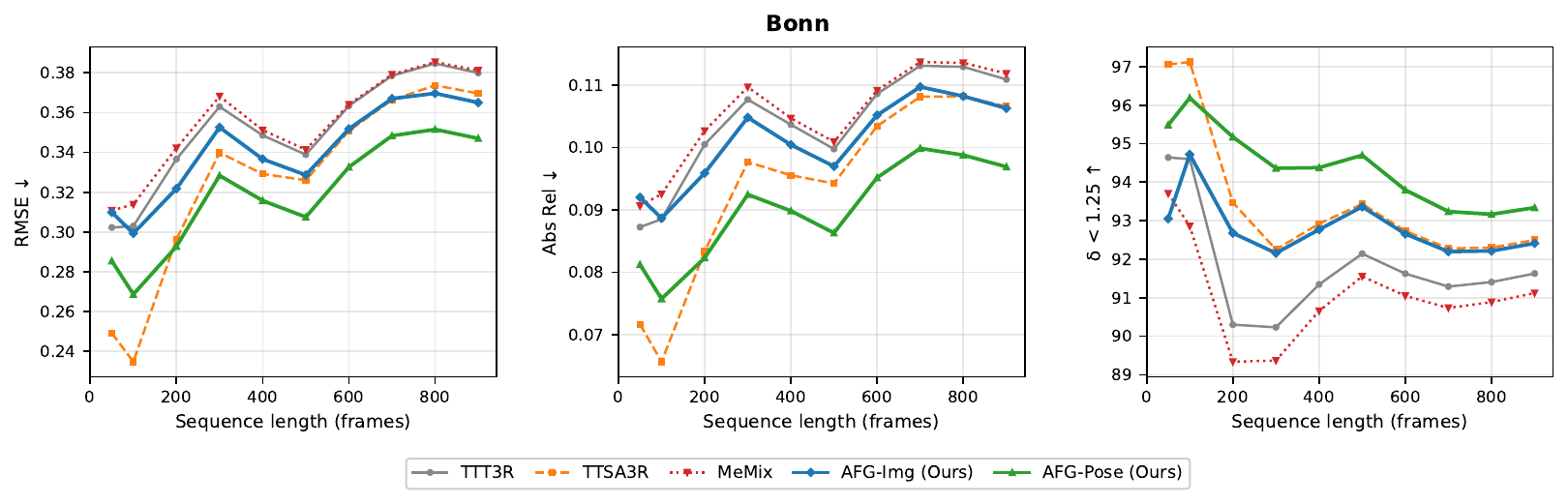}
  \caption{\textbf{Video depth estimation on Bonn (metric alignment).}
  For $L \geq 200$, AFG-Pose improves over TTT3R, TTSA3R, and MeMix on all metrics, with gains widening at longer sequences; at $L \leq 100$ TTSA3R is best.}
  \label{fig:depth_bonn}
\end{figure}

Following prior work~\cite{wang2025continuous,chen2025ttt3r,dong2026memix}, we evaluate video depth estimation on Bonn~\cite{palazzolo2019refusion}, which covers dynamic indoor scenes. We report absolute relative error (Abs Rel), root mean square error (RMSE), and the percentage of predicted depths within a 1.25-factor of ground truth ($\delta < 1.25$) under metric alignment; results are shown in Fig.~\ref{fig:depth_bonn}, with per-length numerical values under both metric and scale-and-shift alignments tabulated in Appendix~\ref{app:depth_extra}. For $L \geq 200$, both AFG-Img and AFG-Pose improve depth estimation over TTT3R and MeMix, with the gains becoming more pronounced on longer sequences where recurrent state degradation accumulates; averaged over all lengths, AFG-Pose reduces AbsRel by $13.0\%$ over TTT3R.
AFG-Pose is the stronger variant throughout, while AFG-Img provides a smaller improvement.
MeMix tracks TTT3R closely at all lengths. On short sequences ($L \leq 100$), where drift has not yet accumulated, TTSA3R is best; AFG-Pose still improves over TTT3R there, while AFG-Img is on par with or slightly behind it.

\subsection{3D Reconstruction}
\label{subsec:exp_recon}

Following common practice in long-horizon streaming reconstruction~\cite{wang2025continuous,chen2025ttt3r,dong2026memix}, we evaluate multiview reconstruction on 7-Scenes~\cite{shotton2013scene} and NRGBD~\cite{azinovic2022neural} at three sequence lengths (300/400/500 frames) with sparse sampling (one frame every two frames, -S); the full per-length, per-baseline table is provided in Table~\ref{tab:recon} of Appendix~\ref{app:recon_quant}.
Full-attention models such as VGGT~\cite{wang2025vggt} run out of memory at all tested lengths.
Among constant-memory recurrent baselines, reconstruction quality degrades consistently as the input horizon grows: TTT3R mean accuracy on 7-Scenes deteriorates from 0.039 at $L{=}300$ to 0.065 at $L{=}500$, and on NRGBD from 0.104 to 0.166, reflecting cumulative state drift under long sequences.

Our AFG variants substantially reduce this length-dependent degradation. On 7-Scenes, AFG-Img maintains near-constant accuracy across all lengths (0.023/0.023/0.024 at $L{=}300/400/500$), while TTT3R degrades by 67\% over the same range.
On NRGBD, where drift is more pronounced, AFG-Img reduces the mean Acc error by 24\% relative to TTSA3R at $L{=}500$ (0.092 vs.\ 0.121) and cuts completeness error by 45\% (0.027 vs.\ 0.049). The MeMix row applies MeMix to TTT3R, the setting proposed in its paper and the like-for-like comparison, since our variants also build on TTT3R. Against the strongest combination reported by MeMix, MeMix on TTSA3R (mean accuracy $0.033$ on 7-Scenes and $0.114$ on NRGBD at $L{=}500$, values from~\cite{dong2026memix}; our TTT3R baseline matches theirs), AFG-Img still reaches $0.024$ and $0.092$. Qualitative comparisons (Fig.~\ref{fig:vis_recon}, Appendix~\ref{app:recon_qual}) corroborate the numbers: without adaptive gating, recurrent baselines accumulate pose drift and produce fragmented geometry, while AFG yields more coherent surfaces and better-preserved scene structure across both datasets.

\subsection{Ablation Studies and Analysis}
\label{subsec:exp_ablation}

\paragraph{Is adaptivity necessary, and which gate carries the gain?}
Table~\ref{tab:ablation_alpha} answers both on TUM-RGBD pose and Bonn depth. \emph{Fixed-$\alpha$ probe.} Replacing the adaptive gate with a constant $\alpha \in \{0.1, \dots, 0.7\}$ already improves ATE over TTT3R, but ATE and rotation are best at opposite ends of the range: ATE is U-shaped with its minimum near $\alpha=0.3$, while rotation error improves monotonically up to $\alpha=0.7$ ($1.849 \to 0.429$). Suppressing drift requires weak writes on redundant frames, whereas fine rotation requires strong writes, and a constant cannot tell the two kinds of frame apart; the adaptive gate improves on the best constant for ATE ($\alpha{=}0.3$) by $18\%$ and on the best constant for AbsRel ($\alpha{=}0.1$, which costs $4\times$ the rotation error) by $3\%$, while keeping rotation error close to TTT3R. \emph{Write-schedule probe.} To test whether the gain comes merely from writing less, we let the state absorb only every $k$-th frame (every frame is still encoded, decoded and evaluated). Writing less helps pose ($0.109\to0.066$ at $k{=}3$) but damages rotation and barely moves depth ($0.098$ vs.\ our $0.086$ AbsRel); on KITTI, writing every third frame gives $53.12$ against our $43.96$. The rest of the gain comes from letting the write strength follow the content (see Appendix~\ref{app:write_less} for subsampled input). \emph{Component probe.} Applying $\alpha$ alone to CUT3R (without $\beta$) already reaches ATE $0.063$ and AbsRel $0.085$, against $0.109$ and $0.100$ for TTT3R, showing the frame-level gate independently captures most of the gain; adding $\beta$ further reduces ATE to $0.054$ ($14\%$ additional) while maintaining depth quality, confirming $\beta$'s complementary token-level selectivity---$\alpha$ controls \emph{when} to update at the frame level, $\beta$ controls \emph{where} at the token level.

\begin{table}[h]
\centering
\caption{\textbf{Ablation: necessity of adaptivity and contribution of each gate.} TUM-RGBD ($1000$ frames, pose) and Bonn ($500$ frames, depth). Blocks sweep a constant $\alpha$, a content-independent write schedule (the state absorbs every $k$-th frame), and the contribution of $\beta$ on top of $\alpha$. ``--'': not measured. Best in \textbf{bold}.}
\label{tab:ablation_alpha}
\setlength{\tabcolsep}{4pt}
\renewcommand{\arraystretch}{0.95}
\footnotesize
\begin{tabular}{l|cc|cc|cc}
\toprule
\multirow{2}{*}{Configuration} & \multirow{2}{*}{$\alpha_t$} & \multirow{2}{*}{$\beta_t$} & \multicolumn{2}{c|}{Pose (TUM)} & \multicolumn{2}{c}{Depth (Bonn)} \\
 & & & ATE$\downarrow$ & RPE$_r$$\downarrow$ & AbsRel$\downarrow$ & $\delta{<}1.25$$\uparrow$ \\
\midrule
TTT3R                                & $\equiv 1$ & \checkmark & 0.109          & 0.443          & 0.100          & 92.1 \\
\midrule
\multicolumn{7}{l}{\textit{Fixed-$\alpha$ probe (does adaptivity matter?)}} \\
$\alpha_t \equiv 0.1$                & $0.1$      & \checkmark & 0.082          & 1.849          & 0.089          & -- \\
$\alpha_t \equiv 0.2$                & $0.2$      & \checkmark & 0.075          & 1.078          & 0.094          & -- \\
$\alpha_t \equiv 0.3$                & $0.3$      & \checkmark & 0.066          & 0.854          & 0.095          & 93.2 \\
$\alpha_t \equiv 0.5$                & $0.5$      & \checkmark & 0.079          & 0.449          & 0.099          & 92.8 \\
$\alpha_t \equiv 0.7$                & $0.7$      & \checkmark & 0.093          & 0.429          & 0.100          & 92.3 \\
\midrule
\multicolumn{7}{l}{\textit{Write-schedule probe (is it only writing less?)}} \\
write every 2nd frame                & $\{0,1\}$  & \checkmark & 0.079          & 0.439          & 0.100          & 91.9 \\
write every 3rd frame                & $\{0,1\}$  & \checkmark & 0.066          & 0.648          & 0.098          & 92.7 \\
\midrule
\multicolumn{7}{l}{\textit{Component probe (does $\beta$ help on top of $\alpha$?)}} \\
$\alpha$ only (no $\beta$)           & adaptive   &            & 0.063          & \textbf{0.376} & \textbf{0.085} & \textbf{95.1} \\
$\alpha + \beta$ (Ours, full)        & adaptive   & \checkmark & \textbf{0.054} & 0.520          & 0.086          & 94.7 \\
\bottomrule
\end{tabular}
\end{table}

\paragraph{Robustness to the gate constants.}
Sweeping $\tau \in [0.5, 1.5]$ for AFG-Pose on TUM-RGBD pose and Bonn depth (Appendix~\ref{app:threshold}), all configurations outperform TTT3R and ATE varies only within $0.051$--$0.064$; sweeping the sharpness $\kappa \in \{1,2,3,5\}$ gives ATE $0.066$--$0.052$ (Appendix~\ref{app:kappa}), so the method does not require per-scene tuning. Larger $\tau$ trades a small RPE$_r$ increase for further ATE and AbsRel gains.

\paragraph{The rotation trade-off.}
At the default setting AFG-Pose raises RPE$_r$ on TUM-RGBD ($0.520$ vs.\ $0.443$) while halving ATE. The cost comes from the setting, not the method: at $\tau=0.5$ it improves both metrics ($0.064$ ATE, $0.376$ RPE$_r$; Appendix~\ref{app:threshold}), and $\alpha$ alone gives the best rotation ($0.376$); we attribute the loss to frames where $\alpha$ and $\beta$ are both low, which receive too little update for fine rotation. The default leans towards ATE.

\paragraph{Does the pose gate track the scene or the state?}
Since $p_t$ is decoded from $S_{t-1}$, $\Delta p_t$ could in principle reflect model instability rather than geometric novelty. On TUM \texttt{walking\_xyz} we compare $\alpha_t$ with two ground-truth quantities that do not depend on our model: the camera rotation since the previous frame, and the pose error at that frame after Sim(3) alignment. The gate follows the camera (Spearman $+0.50$) and not the error ($-0.12$), although the state is never accurate there ($2.9$\,cm accumulated error on average). The coupling is not zero, but it does not drive the gate; moreover, $\alpha_t \leq 1$ means no frame is ever written more strongly than TTT3R would write it, and AFG-Img reads encoder features only and has no such coupling (Appendix~\ref{app:coupling}).

\paragraph{Why not combine the two gates?}
Fusing $\alpha^{\text{img}}$ and $\alpha^{\text{pose}}$ (OR, AND, or a weighted average) never improves over the best single gate, since both draw on correlated per-frame novelty (Appendix~\ref{app:fusion}); we therefore keep them as complementary regime-specific variants.

\section{Conclusion}
\label{conclusion}

In this paper, we rethought why constant-memory recurrent 3D reconstructors drift on long sequences, and traced the failure to a structural blind spot in the per-token gate: averaged across thousands of cross-attention interactions, every gate collapses into nearly the same magnitude, leaving no way to distinguish a novel frame from a redundant one. To restore the missing axis, i.e., frame-level adaptivity, we introduced Adaptive Frame Gating, a parameter-free scalar gate computed at inference time from features the model already produces, with no retraining, no architectural change, and no extra forward pass. Across pose, depth, and reconstruction tasks, AFG outperforms inference-time gating baselines on long sequences, and on long outdoor trajectories surpasses methods purpose-built for long sequences.

\paragraph{Limitations and future work.} AFG closes the frame-level axis but does not address every long-sequence failure mode of fixed-state recurrent models, most notably, state saturation under sustained novelty. Its benefit is tied to accumulated drift: on short sequences it gives little and can be worse (KITTI \texttt{04}; also \texttt{05}, \texttt{07}), since the gate can only protect information that would otherwise have been overwritten. At the default setting it trades some rotation accuracy for lower drift (Sec.~\ref{subsec:exp_ablation}). Because $\alpha_t \leq 1$, AFG only weakens writes; amplifying highly novel frames destabilises the frozen model outdoors (Appendix~\ref{app:cap}).

\begin{ack}
This work was supported in part by the National Natural Science Foundation of China (NSFC) under Grants No.~62525106 and No.~62472046, and in part by the Young Elite Scientists Sponsorship Program of the Beijing High Innovation Plan under Grant No.~20250866. The authors declare no competing interests.
\end{ack}

{
\small
\bibliographystyle{plainnat}
\bibliography{references}
}
\clearpage

\appendix
\section*{Appendix}
\renewcommand{\thesection}{A\arabic{section}}
\setcounter{section}{0}

\section{Redundancy-Injection Probe: Mechanism Verification}
\label{app:redundancy}

We test the central causal claim of the main text---that the structural constancy of $\beta$ produces a short memory horizon, and the resulting long-sequence drift is suppressed by $\alpha$---under a controlled stress test that provides ground-truth zero-information frames and isolates drift from real motion: $100$ pixel-identical frames injected into a real sequence with the GT camera held static.

\paragraph{Setup.}
We take the first $500$ frames of TUM-RGBD \texttt{walking\_xyz} as a normal information stream, then append $100$ pixel-identical copies of frame $499$ at positions $500$--$599$, with the GT camera pose held constant on the injected segment. The same frozen CUT3R checkpoint is run with three update rules: TTT3R (Eq.~\ref{eq:tt3r_beta}), AFG-Img, and AFG-Pose (Eqs.~\ref{eq:img_gate},~\ref{eq:pose_gate}). For each variant we record the per-token gate $\beta_t$, the frame-level gate $\alpha_t$ (where applicable), the per-step state update $\|\Delta S_t\|$, and the estimated camera trajectory. Estimated trajectories are aligned to GT with Sim(3) Umeyama prior to error computation, following the convention of Sec.~\ref{sec:experiments}.

\begin{figure}[h]
\centering
\includegraphics[width=\linewidth]{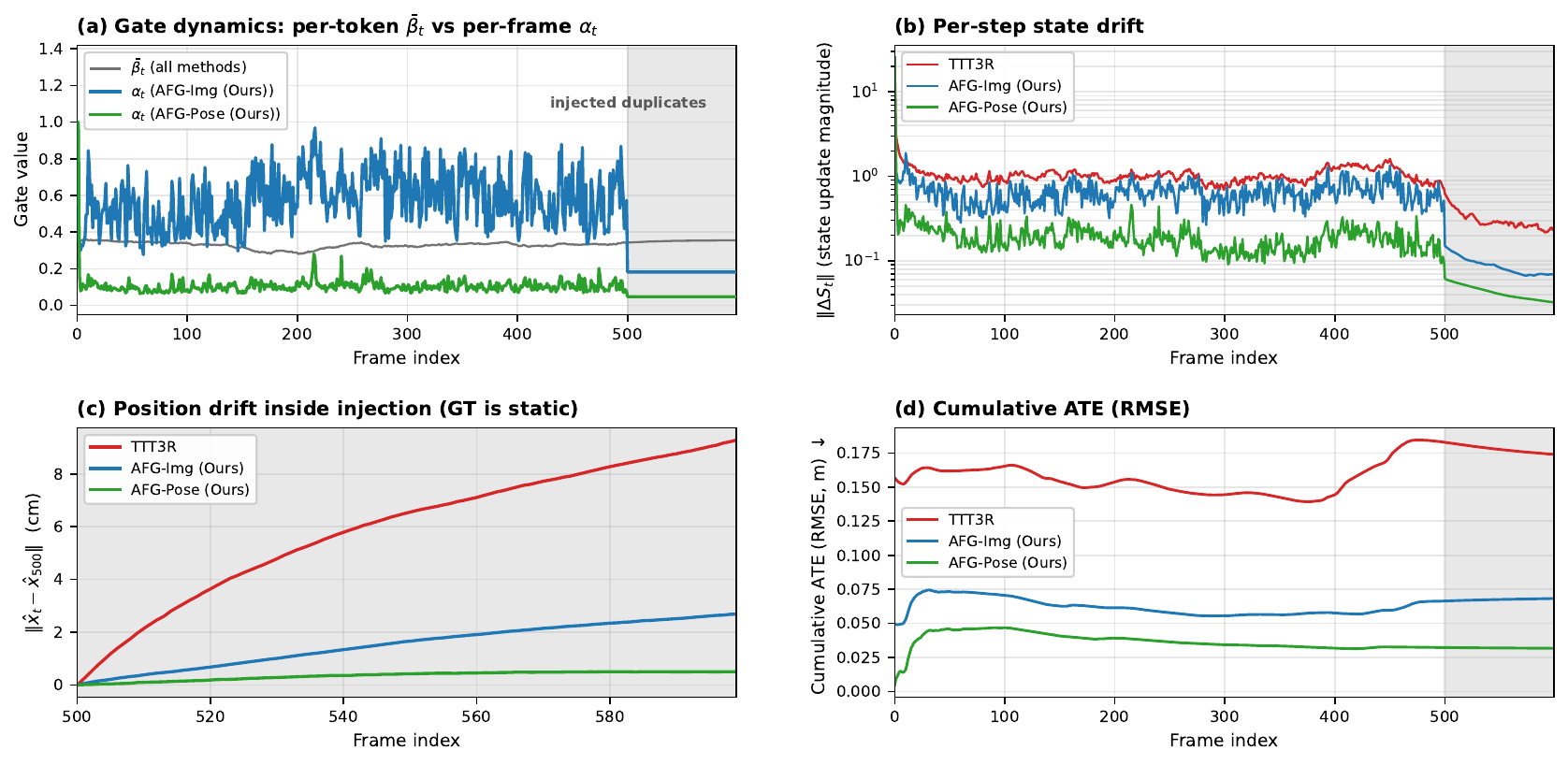}
\caption{\textbf{Redundancy-injection probe on TUM \texttt{walking\_xyz}.} Frames $0$--$499$ are the original sequence; frames $500$--$599$ (gray shading) are $100$ pixel-identical copies of frame $499$ with GT camera held static.
\textbf{(a)} The per-token gate $\bar\beta_t$ (gray; identical across methods) remains flat at $\sim$$0.35$ throughout, including the injected segment, confirming its content-independence; the per-frame gate $\alpha_t$ drops sharply on redundant frames (AFG-Img to $0.18$, AFG-Pose to $0.048$).
\textbf{(b)} The per-step state update $\|\Delta S_t\|$ (log scale): TTT3R retains a non-negligible $\sim$$0.31$ even on identical inputs, while AFG-Pose suppresses it to $0.043$ (a $7\times$ reduction).
\textbf{(c)} Position drift inside the injection, $\|\hat x_t - \hat x_{500}\|$: since GT is static, any non-zero value is pure drift. TTT3R drifts $\sim$$9$\,cm; AFG-Pose stays below $0.5$\,cm ($18\times$ reduction).
\textbf{(d)} Sim(3)-aligned cumulative ATE (RMSE): TTT3R's curve visibly bends upward on the injected segment while AFG-Pose remains flat.
$\bar\beta_t$ and $\alpha_t$ are not comparable in absolute value (different gate types); the diagnostic signal is how each \emph{responds} to redundancy. Colors: TTT3R (\textcolor[HTML]{d62728}{red}), AFG-Img (\textcolor[HTML]{1f77b4}{blue}), AFG-Pose (\textcolor[HTML]{2ca02c}{green}), shared $\bar\beta_t$ (gray).}
\label{fig:redundancy_probe}
\end{figure}

\paragraph{$\beta$ is content-blind, even in the worst case.}
Panel~(a) of Fig.~\ref{fig:redundancy_probe} shows $\bar\beta_t$ holding at $0.352$ on the $100$ injected duplicates, statistically indistinguishable from its value on the preceding informative frames. This is a stronger statement than the dataset-level statistics of Sec.~\ref{subsec:fact1}: even when the input carries strictly zero new information, the per-token gate does not adapt. The architectural origin of bounded $\beta$ identified in Sec.~\ref{subsec:fact1} is therefore not a statistical artifact of typical sequences but a structural property of the gating mechanism itself.

\paragraph{AFG suppresses redundant updates by an order of magnitude.}
Panel~(b) shows that TTT3R's per-step update naturally decreases on identical inputs (because $\Delta S_t$ shrinks when the input residual is small) but plateaus at $\|\Delta S_t\|\approx 0.31$---still large enough to materially perturb the state. AFG-Pose, by additionally gating with $\alpha_t\to 0.048$, drives the effective update down to $0.043$, a further $7\times$ suppression on top of TTT3R's intrinsic decay.

\paragraph{Geometric drift is reduced by an order of magnitude.}
Because the GT camera is held static on $[500, 599]$, the quantity $\|\hat x_t - \hat x_{500}\|$ in panel~(c) is pure drift introduced by the recurrent update. TTT3R drifts to $\sim$$9$\,cm despite no real motion; AFG-Pose stays under $0.5$\,cm---an $18\times$ reduction. Panel~(d) shows the same effect at trajectory scale: TTT3R's cumulative ATE (RMSE) bends upward on the injected segment, while AFG-Pose remains flat.

\paragraph{Numerical closure with the main-text horizon prediction.}
Substituting the measured $\alpha_{\min}=0.048$ and $\bar\beta=0.352$ from panel~(a) into the AFG horizon formula of Sec.~\ref{sec:method} yields
\[
H_{\text{AFG}} \;=\; \frac{1}{\alpha_{\min}\,\bar\beta} \;\approx\; 60 \text{ frames},
\]
while measuring the decay of the drift directly on the injected segment gives $\sim$$64$ frames. Two numbers reached by different routes agree, which makes the mechanism checkable rather than merely plausible. The TTT3R horizon under the same $\bar\beta$ is $1/\bar\beta \approx 2.8$ frames, consistent with the $\sim$3 frames of Sec.~\ref{subsec:fact3}. The structural mechanism diagnosed in Sec.~\ref{sec:analysis} and the remedy proposed in Sec.~\ref{sec:method} are thus both quantitatively confirmed on a single controlled experiment.

\section{Gate Constants and Sensitivity}
\label{app:gate_constants}

Both gates share the form $\alpha_t = \sigma(\kappa(\|\Delta f_t\|_2 - \tau))$ (Eqs.~\ref{eq:img_gate},~\ref{eq:pose_gate}); Table~\ref{tab:gate_constants} lists the constants, fixed across all benchmarks. A sigmoid on a raw distance needs a centre and a width: $\tau$ is the centre, $\kappa$ the sharpness, and the two features live on different scales. The value $\alpha_t \approx 0.09$ shown in Fig.~\ref{fig:teaser} (AFG-Pose) corresponds to $\|\Delta p_t\|_2 \approx 0.23$.

\begin{table}[h]
\centering
\caption{\textbf{Gate constants.} The lowest value is attained at zero feature change.}
\label{tab:gate_constants}
\footnotesize
\begin{tabular}{l|ccc}
\toprule
Variant & Feature & $(\kappa, \tau)$ & Lowest value $\sigma(-\kappa\tau)$ \\
\midrule
AFG-Img & encoder global feature $g_t$ & $(1, 1.5)$ & $0.18$ \\
AFG-Pose & decoder pose token $p_t$ & $(3, 1.0)$ & $0.047$ \\
\bottomrule
\end{tabular}
\end{table}

\subsection{Threshold $\tau$}
\label{app:threshold}

We sweep the threshold $\tau$ of the AFG-Pose gate $\alpha_t = \sigma(\kappa(\|\Delta p_t\|_2 - \tau))$, with $\kappa=3$ fixed, across $\{0.5, 0.75, 1.0, 1.25, 1.5\}$ on TUM-RGBD pose and Bonn depth (Table~\ref{tab:threshold}). Increasing $\tau$ progressively improves ATE and AbsRel at the cost of a higher RPE$_r$, reflecting a trade-off between global trajectory accuracy and local rotational smoothness. All swept configurations significantly outperform TTT3R (ATE = 0.109, AbsRel = 0.100), and ATE varies only within $0.051$--$0.064$, confirming the method does not require per-scene tuning. We adopt $\tau = 1.0$ as the AFG-Pose default in all main-paper experiments. At $\tau=0.5$ AFG-Pose improves over TTT3R on both ATE and RPE$_r$ ($0.064$/$0.376$ vs.\ $0.109$/$0.443$), showing that the rotation cost at the default is a choice of operating point rather than a property of the gate.

\begin{table}[h]
\centering
\caption{\textbf{Threshold sensitivity (AFG-Pose, $\kappa=3$).} Performance varying $\tau$ on TUM-RGBD (1000 frames, pose) and Bonn (500 frames, depth). All configurations outperform TTT3R.}
\label{tab:threshold}
\setlength{\tabcolsep}{4pt}
\renewcommand{\arraystretch}{0.9}
\footnotesize
\begin{tabular}{l|cc|cc}
\toprule
 & \multicolumn{2}{c|}{Pose (TUM)} & \multicolumn{2}{c}{Depth (Bonn)} \\
$\tau$ & ATE$\downarrow$ & RPE$_r$$\downarrow$ & AbsRel$\downarrow$ & $\delta{<}1.25$$\uparrow$ \\
\midrule
0.5 & 0.064 & 0.376 & 0.093 & 94.2 \\
0.75 & 0.058 & 0.391 & 0.090 & 94.7 \\
1.0 (default) & 0.054 & 0.520 & 0.086 & 94.7 \\
1.25 & 0.051 & 0.603 & 0.084 & 94.7 \\
1.5 & 0.058 & 0.751 & 0.082 & 94.7 \\
\bottomrule
\end{tabular}
\end{table}

\subsection{Sharpness $\kappa$}
\label{app:kappa}

Table~\ref{tab:kappa} sweeps $\kappa$ for AFG-Pose with $\tau=1.0$. Every setting improves substantially over TTT3R; even $\kappa=1$, i.e.\ the gate without a sharpness constant, reduces ATE from $0.109$ to $0.066$ and RPE$_r$ from $0.443$ to $0.421$. Larger $\kappa$ trades rotation for drift, mirroring the $\tau$ sweep above.

\begin{table}[h]
\centering
\caption{\textbf{Sharpness sweep (AFG-Pose, $\tau=1.0$).} TUM-RGBD ($1000$ frames) and Bonn ($500$ frames). ``--'': not measured.}
\label{tab:kappa}
\footnotesize
\begin{tabular}{l|cc|c}
\toprule
$\kappa$ & TUM ATE$\downarrow$ & TUM RPE$_r$$\downarrow$ & Bonn AbsRel$\downarrow$ \\
\midrule
TTT3R (no frame gate) & 0.109 & 0.443 & 0.100 \\
\midrule
1 & 0.066 & 0.421 & 0.095 \\
2 & 0.059 & -- & -- \\
3 (default) & 0.054 & 0.520 & 0.086 \\
5 & \textbf{0.052} & -- & -- \\
\bottomrule
\end{tabular}
\end{table}

\section{Capping the Frame Gate at 1}
\label{app:cap}

Since $\alpha_t$ acts as an inference-time learning rate, values above $1$ could in principle help highly novel frames. We amplify only the frames the gate already treats as novel: $\alpha_t = 1$ is remapped to a cap $s$, while frames at or below the threshold are left untouched. Indoor results use the protocol of Table~\ref{tab:ablation_alpha}; KITTI uses all 11 sequences under the protocol of Table~\ref{tab:kitti}.

\begin{table}[h]
\centering
\caption{\textbf{Raising the cap of the frame gate (AFG-Pose).}}
\label{tab:cap}
\footnotesize
\begin{tabular}{l|cc|c|c}
\toprule
Cap $s$ & TUM ATE$\downarrow$ & TUM RPE$_r$$\downarrow$ & Bonn AbsRel$\downarrow$ & KITTI ATE$\downarrow$ (mean of 11) \\
\midrule
TTT3R (no frame gate) & 0.109 & 0.443 & 0.100 & 68.54 \\
\midrule
$s=1$ (Ours) & \textbf{0.054} & 0.520 & 0.086 & \textbf{43.96} \\
$s=2$ & 0.055 & 0.478 & 0.083 & 55.88 \\
$s=5$ & 0.058 & \textbf{0.411} & \textbf{0.078} & 97.71 \\
\bottomrule
\end{tabular}
\end{table}

Indoors, raising the cap changes little: ATE rises slightly while rotation and depth improve. On KITTI the error rises steadily with the cap; at $s=5$ it is worse than using no frame gate at all on 9 of the 11 sequences, and on sequence \texttt{09} the capped gate is $64\%$ better than TTT3R while $s=5$ is $583\%$ worse. The difference lies in how the gate values are distributed: indoors only a few percent of frames clear the threshold, so amplifying them touches almost nothing, whereas in fast outdoor motion a much larger share clears it, and updates larger than anything seen during training accumulate until the state degrades. Scaling the whole gate, which raises its minimum as well as its cap, is worse still (TUM ATE $0.054 \to 0.060$ for AFG-Pose, $0.084 \to 0.112$ for AFG-Img). The cap therefore keeps AFG safe on a frozen checkpoint: its worst case is exactly TTT3R's own update.

\section{Comparison with Keyframe-Style Gating}
\label{app:keyframe}

To separate AFG from keyframe selection at the level of mechanism (Sec.~\ref{subsec:kf_relation}), we vary one choice at a time while keeping the feature and threshold of our gate. \emph{Hard vs.\ graded}: we binarise the gate, $\alpha_t = 1$ above the threshold and a fixed low value otherwise. \emph{Dropping vs.\ keeping}: the low value is set to $0.05$ (every frame keeps writing weakly) or to $0$ (the classical admit/skip rule). \emph{Reference frame}: the feature change is measured against the last accepted frame (reset whenever the gate opens) instead of the previous frame, closer to classical keyframe criteria. All runs use the frozen checkpoint and the protocol of Table~\ref{tab:ablation_alpha}.

\begin{table}[h]
\centering
\caption{\textbf{Keyframe-style variants of the frame gate.} TUM-RGBD ($1000$ frames), ScanNet ($1000$ frames), Bonn ($500$ frames). ``--'': not measured.}
\label{tab:keyframe}
\footnotesize
\begin{tabular}{l|cc|cc|c}
\toprule
 & \multicolumn{2}{c|}{TUM (AFG-Pose)} & \multicolumn{2}{c|}{ScanNet ATE$\downarrow$} & Bonn \\
Variant & ATE$\downarrow$ & RPE$_r$$\downarrow$ & AFG-Img & AFG-Pose & AbsRel$\downarrow$ \\
\midrule
TTT3R (no frame gate) & 0.109 & 0.443 & \multicolumn{2}{c|}{0.405} & 0.100 \\
\midrule
binarised, low value $0$ (admit/skip) & 0.061 & 0.880 & -- & -- & -- \\
binarised, low value $0.05$ & \textbf{0.052} & 0.647 & 0.300 & 0.352 & -- \\
graded, last accepted frame as reference & 0.065 & -- & -- & -- & 0.092 \\
graded, previous frame (Ours) & 0.054 & \textbf{0.520} & \textbf{0.282} & \textbf{0.281} & \textbf{0.086} \\
\bottomrule
\end{tabular}
\end{table}

Dropping non-keyframes is the most harmful choice: a zero-weight frame can only be recovered by later frames writing more, whereas with a positive minimum an under-rated frame is merely delayed. On TUM the hard and graded responses tie on ATE but the graded one is better on rotation, and on ScanNet the graded gate is clearly better for both variants. Measuring novelty against the last accepted frame is worse on both pose and depth, so the adjacent-frame change is not only the simpler signal but the better one for this update rule.

\section{Is the Gain Only from Writing Less?}
\label{app:write_less}

\paragraph{Write schedule on full-rate input.} Table~\ref{tab:ablation_alpha} compares AFG against content-independent write schedules, in which every frame is still encoded, decoded and evaluated but the state absorbs only every $k$-th frame. Changing only how often the state is written gives, on KITTI (mean ATE over 11 sequences), $68.54$ when writing every frame (TTT3R), $53.12$ when writing every third frame, and $43.96$ when the content sets the write strength (AFG-Pose). Writing less helps pose both indoors and outdoors, but the gate lowers the error by a further $18\%$ on TUM ($0.066 \to 0.054$) and $17\%$ on KITTI, and on depth nearly all of the gain comes from content (Table~\ref{tab:ablation_alpha}).

\paragraph{Subsampled input.} We additionally subsample the \emph{input} stream, feeding only every third frame ($k=3$) to both TTT3R and AFG-Pose, so that both models see exactly the same frames and are scored at exactly the same points; the ScanNet benchmark is already built on a stride-3 stream. Table~\ref{tab:input_stride} shows that on subsampled input the frame gate still improves depth, KITTI and ScanNet, but is worse on TUM (ATE $+8\%$, RPE$_r$ $+22\%$). Subsampling and gating act on different quantities (how many frames enter the model vs.\ how strongly each writes into the state), and their gains combine on most benchmarks but not on all. On the same ScanNet stream, the per-token methods TTSA3R ($0.388$) and MeMix ($0.367$) stay close to TTT3R, so a fixed stride combined with token-level adaptivity does not replace frame-level control.

\begin{table}[h]
\centering
\caption{\textbf{Input subsampling (stride $k=3$), same frames for both methods.} ScanNet is natively evaluated on a stride-3 stream.}
\label{tab:input_stride}
\footnotesize
\begin{tabular}{l|cc}
\toprule
Benchmark / metric & TTT3R & AFG-Pose \\
\midrule
TUM, ATE$\downarrow$ & \textbf{0.0604} & 0.0650 \\
TUM, RPE$_r$$\downarrow$ & \textbf{0.934} & 1.137 \\
Bonn, AbsRel$\downarrow$ & 0.0978 & \textbf{0.0861} \\
Bonn, $\delta{<}1.25$$\uparrow$ & 92.5 & \textbf{94.3} \\
KITTI, ATE$\downarrow$ (mean of 11) & 106.79 & \textbf{95.69} \\
ScanNet, ATE$\downarrow$ & 0.405 & \textbf{0.281} \\
\bottomrule
\end{tabular}
\end{table}

The gain on subsampled input also grows with length. On the four longest KITTI sequences at stride $3$ (Table~\ref{tab:stride_kitti_long}), AFG-Pose lowers the mean ATE by $21.6\%$.

\begin{table}[h]
\centering
\caption{\textbf{Four longest KITTI sequences at input stride $k=3$.} ATE$\downarrow$ (m).}
\label{tab:stride_kitti_long}
\footnotesize
\begin{tabular}{c|c|cc}
\toprule
Sequence & Frames & TTT3R & AFG-Pose \\
\midrule
02 & 4661 & 241.25 & \textbf{208.59} \\
00 & 4541 & 178.61 & \textbf{117.74} \\
08 & 4071 & 168.44 & \textbf{117.21} \\
05 & 2761 & 121.37 & \textbf{112.97} \\
\midrule
Mean & & 177.42 & \textbf{139.13} ($-21.6\%$) \\
\bottomrule
\end{tabular}
\end{table}

\section{Coupling Between the Pose Gate and the Recurrent State}
\label{app:coupling}

The pose token $p_t$ is decoded from the current image and the state $S_{t-1}$, so $\Delta p_t$ measures how much a frame changes the model's pose inference, and could in principle reflect a drifting state rather than geometric novelty. If the gate were driven by the model's own error, it should open widest where that error is largest. On TUM \texttt{walking\_xyz} we record $\alpha_t$ for every frame together with two ground-truth quantities that do not rely on our model: the camera rotation since the previous frame, and the pose error at that frame (position gap after Sim(3) alignment).

\begin{itemize}
    \item \textbf{Correlation.} Over all frames, $\alpha_t$ has a Spearman rank correlation of $+0.50$ with the camera rotation and $-0.12$ with the pose error. Rank correlation is used because the gate is heavy-tailed.
    \item \textbf{Extremes.} The top $10\%$ of frames by $\alpha_t$ carry $1.68\times$ the camera rotation of the bottom $50\%$ ($0.82^\circ$ vs.\ $0.49^\circ$ per frame), while their pose error is level, if anything lower ($2.92$ vs.\ $3.07$\,cm).
    \item \textbf{The state is not accurate.} The accumulated pose error on this sequence is $2.9$\,cm on average and $6.6$\,cm at worst, so the absence of a link is not due to an error-free state.
\end{itemize}

Three qualifications apply. (i) We do not claim the coupling is zero: that $p_t$ reads the state is an architectural fact; our claim is that it does not drive the gate. (ii) Whether the state accumulates error is a property of the frozen backbone; AFG only rescales each frame's write, and with $\alpha_t \leq 1$ no frame is ever written more strongly than TTT3R would write it, whatever the state. (iii) AFG-Img reads encoder features only and never touches the state, so this coupling has no path there; the two variants nevertheless agree in trend on all six benchmarks.

\section{Combining the Two Gates}
\label{app:fusion}

A natural question is whether AFG-Img and AFG-Pose can be combined. We evaluate three fusion strategies in Table~\ref{tab:combine_variants}: OR via $\max(\alpha^{\text{img}}, \alpha^{\text{pose}})$; multiplicative AND $\alpha^{\text{img}}\cdot\alpha^{\text{pose}}$ with relaxed thresholds; and a signal-strength weighted average. No fusion strategy improves over the best single gate. The OR variant lets too many redundant frames through (TUM ATE $0.079$ vs.\ AFG-Pose $0.054$), undoing selectivity. The multiplicative variant cuts ATE moderately (0.075) but explodes the rotation error (RPE$_r$ $1.137$, $2.6\times$ TTT3R) by zeroing informative residuals when both gates fire low. The weighted variant matches AFG-Pose within noise (ATE $0.054$, AbsRel $0.086$): both gates draw from the same per-frame backbone activations and carry correlated novelty, so fusion adds no new selectivity. We therefore adopt the two as \emph{complementary regime-specific variants} rather than fusing them.

\begin{table}[h]
\centering
\caption{\textbf{Combining the two gates is not beneficial.} Fusion strategies for $\alpha^{\text{img}}$ and $\alpha^{\text{pose}}$ on TUM-RGBD ($1000$ frames, pose) and Bonn ($500$ frames, depth). No combination strictly improves over the best single gate. Best in \textbf{bold}.}
\label{tab:combine_variants}
\setlength{\tabcolsep}{4pt}
\renewcommand{\arraystretch}{0.9}
\footnotesize
\begin{tabular}{l|cc|ccc}
\toprule
\multirow{2}{*}{Configuration} & \multicolumn{2}{c|}{Pose (TUM)} & \multicolumn{3}{c}{Depth (Bonn)} \\
 & ATE$\downarrow$ & RPE$_r$$\downarrow$ & AbsRel$\downarrow$ & RMSE$\downarrow$ & $\delta{<}1.25$$\uparrow$ \\
\midrule
TTT3R                                                         & 0.109 & 0.443 & 0.100 & 0.339 & 92.1 \\
\midrule
AFG-Img alone                                                 & 0.084 & \textbf{0.411} & 0.097 & 0.329 & 93.4 \\
AFG-Pose alone                                                & \textbf{0.054} & 0.520 & \textbf{0.086} & \textbf{0.308} & \textbf{94.7} \\
\midrule
$\max(\alpha^{\text{img}}, \alpha^{\text{pose}})$ \,(OR)            & 0.079 & 0.414 & 0.095 & 0.323 & 94.0 \\
$\alpha^{\text{img}}\cdot\alpha^{\text{pose}}$ \,(retuned)          & 0.075 & 1.137 & 0.089 & 0.315 & 94.1 \\
weighted$(\alpha^{\text{img}}, \alpha^{\text{pose}})$              & \textbf{0.054} & 0.522 & \textbf{0.086} & \textbf{0.308} & \textbf{94.7} \\
\bottomrule
\end{tabular}
\end{table}

\section{Detailed Per-Length Pose Numbers}
\label{app:pose_detailed}

We report the per-length numerical values underlying Fig.~\ref{fig:pose_ate} of Sec.~\ref{subsec:exp_pose}. Tables~\ref{tab:tum_pose_full} and~\ref{tab:scannet_pose_full} list ATE on TUM-RGBD and ScanNet across sequence lengths from $50$ to $1000$ frames, for TTT3R~\cite{chen2025ttt3r}, TTSA3R~\cite{zheng2026ttsa3r}, MeMix~\cite{dong2026memix}, and our two variants.

\begin{table}[h]
\centering
\caption{\textbf{Detailed pose ATE on TUM-RGBD across sequence lengths.} ATE $\downarrow$ (m) under Sim(3) Umeyama alignment; values averaged over sequences. Best in \textbf{bold}.}
\label{tab:tum_pose_full}
\setlength{\tabcolsep}{4pt}
\renewcommand{\arraystretch}{0.9}
\footnotesize
\begin{tabular}{l|cccccccccc}
\toprule
\multirow{2}{*}{Method} & \multicolumn{10}{c}{Sequence Length $L$} \\
\cmidrule(lr){2-11}
 & 50 & 100 & 200 & 300 & 400 & 500 & 600 & 700 & 800 & 1000 \\
\midrule
TTT3R~\cite{chen2025ttt3r}        & 0.014 & 0.020 & 0.041 & 0.045 & 0.051 & 0.067 & 0.074 & 0.085 & 0.091 & 0.109 \\
TTSA3R~\cite{zheng2026ttsa3r}     & 0.013 & \textbf{0.017} & 0.030 & 0.038 & 0.043 & 0.056 & 0.062 & 0.070 & 0.076 & 0.093 \\
MeMix~\cite{dong2026memix}        & 0.014 & 0.020 & 0.033 & 0.042 & 0.047 & 0.060 & 0.068 & 0.074 & 0.085 & 0.101 \\
\midrule
AFG-Img \textbf{(Ours)}           & \textbf{0.012} & \textbf{0.017} & 0.027 & 0.033 & 0.037 & 0.044 & 0.051 & 0.058 & 0.066 & 0.084 \\
AFG-Pose \textbf{(Ours)}          & 0.013 & \textbf{0.017} & \textbf{0.023} & \textbf{0.028} & \textbf{0.032} & \textbf{0.033} & \textbf{0.036} & \textbf{0.040} & \textbf{0.045} & \textbf{0.054} \\
\bottomrule
\end{tabular}
\end{table}

\begin{table}[h]
\centering
\caption{\textbf{Detailed pose ATE on ScanNet across sequence lengths.} ATE $\downarrow$ (m) under Sim(3) Umeyama alignment. Best in \textbf{bold}.}
\label{tab:scannet_pose_full}
\setlength{\tabcolsep}{4pt}
\renewcommand{\arraystretch}{0.9}
\footnotesize
\begin{tabular}{l|cccccccccc}
\toprule
\multirow{2}{*}{Method} & \multicolumn{10}{c}{Sequence Length $L$} \\
\cmidrule(lr){2-11}
 & 50 & 100 & 200 & 300 & 400 & 500 & 600 & 700 & 800 & 1000 \\
\midrule
TTT3R~\cite{chen2025ttt3r}        & 0.033 & 0.072 & 0.144 & 0.192 & 0.236 & 0.278 & 0.315 & 0.345 & 0.371 & 0.405 \\
TTSA3R~\cite{zheng2026ttsa3r}     & \textbf{0.032} & \textbf{0.063} & \textbf{0.122} & 0.169 & 0.213 & 0.259 & 0.298 & 0.329 & 0.356 & 0.388 \\
MeMix~\cite{dong2026memix}        & 0.036 & 0.075 & 0.147 & 0.194 & 0.231 & 0.261 & 0.288 & 0.308 & 0.330 & 0.367 \\
\midrule
AFG-Img \textbf{(Ours)}           & 0.033 & 0.071 & 0.135 & \textbf{0.165} & \textbf{0.194} & \textbf{0.212} & \textbf{0.231} & \textbf{0.249} & \textbf{0.265} & 0.282 \\
AFG-Pose \textbf{(Ours)}          & 0.038 & 0.084 & 0.173 & 0.206 & 0.240 & 0.250 & 0.257 & 0.267 & 0.277 & \textbf{0.281} \\
\bottomrule
\end{tabular}
\end{table}

\section{Per-Sequence Gains on KITTI}
\label{app:kitti_length}

Table~\ref{tab:kitti_length} re-orders the KITTI results of Table~\ref{tab:kitti} by sequence length and reports the relative change of AFG-Pose over TTT3R. The largest improvements occur on the longest sequences (all three sequences above $4000$ frames improve by $39$--$51\%$), and the shortest sequence (\texttt{04}) gets worse. The relation is a trend rather than a rule: \texttt{05} and \texttt{07} degrade despite their length. AFG has no loop-closure component, and loop structure does not separate the outcomes (Sec.~\ref{subsec:exp_kitti}). The gate can only protect information that would otherwise have been overwritten, and on a short trajectory little has been; we list this as a limitation.

\begin{table}[h]
\centering
\caption{\textbf{KITTI ATE (m) ordered by sequence length.} Change is relative to TTT3R; negative is better.}
\label{tab:kitti_length}
\footnotesize
\begin{tabular}{c|c|c|cc|cc}
\toprule
Seq. & Frames & TTT3R & AFG-Pose & Change & AFG-Img & Change \\
\midrule
02 & 4661 & 242.05 & 118.98 & $-50.8\%$ & 229.25 & $-5.3\%$ \\
00 & 4541 & 118.20 & 72.22 & $-38.9\%$ & 89.03 & $-24.7\%$ \\
08 & 4071 & 81.52 & 45.92 & $-43.7\%$ & 49.61 & $-39.1\%$ \\
05 & 2761 & 31.62 & 35.16 & $+11.2\%$ & 32.21 & $+1.9\%$ \\
09 & 1591 & 79.10 & 28.56 & $-63.9\%$ & 47.21 & $-40.3\%$ \\
10 & 1201 & 29.82 & 31.09 & $+4.3\%$ & 27.44 & $-8.0\%$ \\
01 & 1101 & 102.98 & 85.49 & $-17.0\%$ & 79.40 & $-22.9\%$ \\
06 & 1101 & 37.77 & 30.15 & $-20.2\%$ & 31.29 & $-17.2\%$ \\
07 & 1101 & 12.00 & 17.44 & $+45.3\%$ & 14.01 & $+16.8\%$ \\
03 & 801 & 14.38 & 12.07 & $-16.1\%$ & 14.68 & $+2.1\%$ \\
04 & 271 & 4.51 & 6.49 & $+43.9\%$ & 8.47 & $+87.8\%$ \\
\bottomrule
\end{tabular}
\end{table}

\section{KITTI Trajectory Visualizations}
\label{app:kitti_traj}

\begin{figure}[p]
\centering
\includegraphics[width=\linewidth]{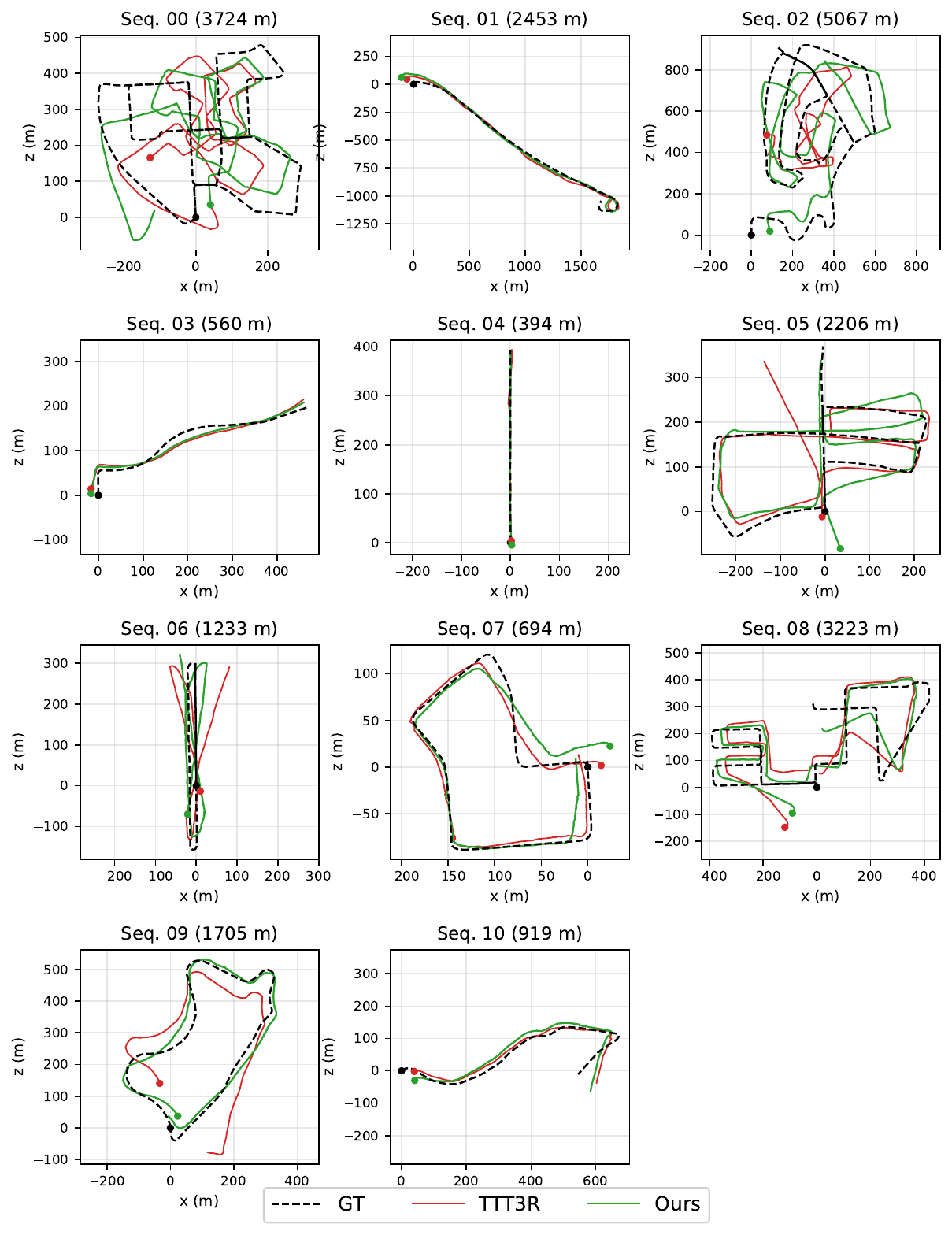}
\caption{\textbf{Camera trajectory visualizations on KITTI Odometry.} Top-down ($x$--$z$) views of the recovered trajectories on all 11 sequences (\texttt{00}--\texttt{10}). Each estimated trajectory is aligned to ground truth via Sim(3) Umeyama; panel titles list the official KITTI path length in meters. On most long sequences ($\geq$ 1200~m), TTT3R~\cite{chen2025ttt3r} drifts substantially while Ours (AFG-Pose) tracks the ground-truth route more closely. On short sequences ($\leq$ 920~m), both methods track ground truth accurately, consistent with our finding that the gating advantage strengthens as accumulated drift grows.}
\label{fig:kitti_traj}
\end{figure}

Fig.~\ref{fig:kitti_traj} visualizes the recovered camera trajectories on all $11$ KITTI Odometry sequences (\texttt{00}--\texttt{10}) for ground truth, TTT3R~\cite{chen2025ttt3r}, and our AFG-Pose variant (labeled ``Ours''), corresponding to the per-sequence ATE numbers in Table~\ref{tab:kitti}. Each estimated trajectory is aligned to ground truth by Sim(3) Umeyama; panel titles report the official KITTI path length. On most long sequences (\texttt{00}, \texttt{01}, \texttt{02}, \texttt{06}, \texttt{08}, \texttt{09}; $\geq$ 1200 m), TTT3R drifts visibly while Ours tracks the ground-truth route more closely, directly visualizing the long-sequence drift suppression; on \texttt{05} the two are comparable (ATE $35.16$ vs.\ $31.62$). On short sequences (\texttt{03}, \texttt{04}, \texttt{07}, \texttt{10}; $\leq$ 920 m), both methods track ground truth closely, consistent with the observation that the gating advantage emerges as accumulated drift grows.

\section{Detailed Bonn Depth Numbers}
\label{app:depth_extra}

We provide the per-length numerical values plotted in Fig.~\ref{fig:depth_bonn} of Sec.~\ref{subsec:exp_depth} under metric alignment (Sec.~\ref{app:bonn_metric}), and additionally report results under scale-and-shift alignment (Sec.~\ref{app:bonn_ss}) as a robustness check, since scale-and-shift is the standard alignment in the depth-estimation literature. Across both alignments and all three metrics, AFG-Pose is the strongest variant for $L \geq 200$; TTSA3R wins at very short sequences ($L \leq 100$) where short-horizon visual matching dominates and drift has not yet accumulated.

\subsection{Metric Alignment}
\label{app:bonn_metric}

Tables~\ref{tab:bonn_metric_absrel}, \ref{tab:bonn_metric_rmse}, and~\ref{tab:bonn_metric_d125} list AbsRel, RMSE, and $\delta{<}1.25$ under metric alignment, the strictest evaluation (no scale or shift compensation), corresponding directly to the curves of Fig.~\ref{fig:depth_bonn}.

\begin{table}[h]
\centering
\caption{\textbf{Bonn depth (Abs Rel $\downarrow$) under metric alignment, across sequence lengths.} Best in \textbf{bold}.}
\label{tab:bonn_metric_absrel}
\setlength{\tabcolsep}{4pt}
\renewcommand{\arraystretch}{0.9}
\footnotesize
\begin{tabular}{l|cccccccccc}
\toprule
\multirow{2}{*}{Method} & \multicolumn{10}{c}{Sequence Length $L$} \\
\cmidrule(lr){2-11}
 & 50 & 100 & 200 & 300 & 400 & 500 & 600 & 700 & 800 & 900 \\
\midrule
TTT3R~\cite{chen2025ttt3r}        & 0.0872 & 0.0885 & 0.1005 & 0.1077 & 0.1036 & 0.0997 & 0.1086 & 0.1131 & 0.1129 & 0.1109 \\
TTSA3R~\cite{zheng2026ttsa3r}     & \textbf{0.0717} & \textbf{0.0657} & 0.0834 & 0.0977 & 0.0955 & 0.0942 & 0.1034 & 0.1081 & 0.1082 & 0.1066 \\
MeMix~\cite{dong2026memix}        & 0.0906 & 0.0925 & 0.1027 & 0.1097 & 0.1047 & 0.1009 & 0.1091 & 0.1137 & 0.1135 & 0.1118 \\
\midrule
AFG-Img \textbf{(Ours)}           & 0.0920 & 0.0887 & 0.0959 & 0.1048 & 0.1004 & 0.0970 & 0.1052 & 0.1097 & 0.1082 & 0.1063 \\
AFG-Pose \textbf{(Ours)}          & 0.0812 & 0.0758 & \textbf{0.0823} & \textbf{0.0925} & \textbf{0.0898} & \textbf{0.0863} & \textbf{0.0952} & \textbf{0.0998} & \textbf{0.0988} & \textbf{0.0969} \\
\bottomrule
\end{tabular}
\end{table}

\begin{table}[h]
\centering
\caption{\textbf{Bonn depth (RMSE $\downarrow$) under metric alignment, across sequence lengths.} Best in \textbf{bold}.}
\label{tab:bonn_metric_rmse}
\setlength{\tabcolsep}{4pt}
\renewcommand{\arraystretch}{0.9}
\footnotesize
\begin{tabular}{l|cccccccccc}
\toprule
\multirow{2}{*}{Method} & \multicolumn{10}{c}{Sequence Length $L$} \\
\cmidrule(lr){2-11}
 & 50 & 100 & 200 & 300 & 400 & 500 & 600 & 700 & 800 & 900 \\
\midrule
TTT3R~\cite{chen2025ttt3r}        & 0.302 & 0.303 & 0.337 & 0.363 & 0.349 & 0.339 & 0.363 & 0.379 & 0.385 & 0.380 \\
TTSA3R~\cite{zheng2026ttsa3r}     & \textbf{0.249} & \textbf{0.235} & 0.296 & 0.340 & 0.329 & 0.326 & 0.351 & 0.366 & 0.374 & 0.370 \\
MeMix~\cite{dong2026memix}        & 0.311 & 0.314 & 0.342 & 0.368 & 0.351 & 0.341 & 0.364 & 0.379 & 0.385 & 0.381 \\
\midrule
AFG-Img \textbf{(Ours)}           & 0.310 & 0.299 & 0.322 & 0.353 & 0.337 & 0.329 & 0.352 & 0.367 & 0.370 & 0.365 \\
AFG-Pose \textbf{(Ours)}          & 0.285 & 0.269 & \textbf{0.293} & \textbf{0.328} & \textbf{0.316} & \textbf{0.308} & \textbf{0.333} & \textbf{0.348} & \textbf{0.352} & \textbf{0.347} \\
\bottomrule
\end{tabular}
\end{table}

\begin{table}[h]
\centering
\caption{\textbf{Bonn depth ($\delta{<}1.25$ $\uparrow$, \%) under metric alignment, across sequence lengths.} Best in \textbf{bold}.}
\label{tab:bonn_metric_d125}
\setlength{\tabcolsep}{4pt}
\renewcommand{\arraystretch}{0.9}
\footnotesize
\begin{tabular}{l|cccccccccc}
\toprule
\multirow{2}{*}{Method} & \multicolumn{10}{c}{Sequence Length $L$} \\
\cmidrule(lr){2-11}
 & 50 & 100 & 200 & 300 & 400 & 500 & 600 & 700 & 800 & 900 \\
\midrule
TTT3R~\cite{chen2025ttt3r}        & 94.6 & 94.6 & 90.3 & 90.2 & 91.3 & 92.1 & 91.6 & 91.3 & 91.4 & 91.6 \\
TTSA3R~\cite{zheng2026ttsa3r}     & \textbf{97.1} & \textbf{97.1} & 93.5 & 92.3 & 92.9 & 93.4 & 92.7 & 92.3 & 92.3 & 92.5 \\
MeMix~\cite{dong2026memix}        & 93.7 & 92.9 & 89.3 & 89.4 & 90.7 & 91.6 & 91.0 & 90.7 & 90.9 & 91.1 \\
\midrule
AFG-Img \textbf{(Ours)}           & 93.1 & 94.7 & 92.7 & 92.2 & 92.8 & 93.4 & 92.7 & 92.2 & 92.2 & 92.4 \\
AFG-Pose \textbf{(Ours)}          & 95.5 & 96.2 & \textbf{95.2} & \textbf{94.4} & \textbf{94.4} & \textbf{94.7} & \textbf{93.8} & \textbf{93.2} & \textbf{93.2} & \textbf{93.3} \\
\bottomrule
\end{tabular}
\end{table}

\subsection{Scale-and-Shift Alignment}
\label{app:bonn_ss}

Table~\ref{tab:bonn_ss_full} reports AbsRel under scale-and-shift alignment, the standard convention where each predicted depth map is fitted to ground truth via a per-frame scale and shift before evaluation. Patterns are consistent with the metric-alignment results above, confirming that the gains are not artifacts of the alignment protocol.

\begin{table}[h]
\centering
\caption{\textbf{Bonn depth (Abs Rel $\downarrow$) under scale-and-shift alignment, across sequence lengths.} Best in \textbf{bold}.}
\label{tab:bonn_ss_full}
\setlength{\tabcolsep}{4pt}
\renewcommand{\arraystretch}{0.9}
\footnotesize
\begin{tabular}{l|cccccccccc}
\toprule
\multirow{2}{*}{Method} & \multicolumn{10}{c}{Sequence Length $L$} \\
\cmidrule(lr){2-11}
 & 50 & 100 & 200 & 300 & 400 & 500 & 600 & 700 & 800 & 900 \\
\midrule
TTT3R~\cite{chen2025ttt3r}        & 0.0699 & 0.0608 & 0.0636 & 0.0751 & 0.0747 & 0.0720 & 0.0833 & 0.0899 & 0.0909 & 0.0894 \\
TTSA3R~\cite{zheng2026ttsa3r}     & \textbf{0.0575} & \textbf{0.0524} & 0.0684 & 0.0779 & 0.0765 & 0.0749 & 0.0854 & 0.0917 & 0.0926 & 0.0912 \\
MeMix~\cite{dong2026memix}        & 0.0716 & 0.0624 & 0.0630 & 0.0749 & 0.0746 & 0.0721 & 0.0827 & 0.0894 & 0.0902 & 0.0888 \\
\midrule
AFG-Img \textbf{(Ours)}           & 0.0646 & 0.0557 & 0.0593 & 0.0691 & 0.0683 & 0.0665 & 0.0770 & 0.0841 & 0.0842 & 0.0829 \\
AFG-Pose \textbf{(Ours)}          & 0.0618 & 0.0542 & \textbf{0.0585} & \textbf{0.0675} & \textbf{0.0674} & \textbf{0.0654} & \textbf{0.0758} & \textbf{0.0821} & \textbf{0.0821} & \textbf{0.0808} \\
\bottomrule
\end{tabular}
\end{table}

\section{Detailed 3D Reconstruction Numbers}
\label{app:recon_quant}

We provide the per-length, per-baseline reconstruction metrics on 7-Scenes~\cite{shotton2013scene} and NRGBD~\cite{azinovic2022neural} that underlie the analysis in Sec.~\ref{subsec:exp_recon}. Table~\ref{tab:recon} reports Accuracy, Completeness, and Normal Consistency (mean and median) at three sequence lengths $L \in \{300, 400, 500\}$, comparing TTT3R~\cite{chen2025ttt3r}, MeMix~\cite{dong2026memix}, TTSA3R~\cite{zheng2026ttsa3r}, and our two variants; full-attention VGGT~\cite{wang2025vggt} runs out of memory at all tested lengths. Both AFG variants substantially improve over the inference-time baselines, with the gap widening at longer sequences.

\begin{table*}[h]
\centering
\caption{%
  \textbf{3D reconstruction on 7-Scenes~\cite{shotton2013scene} and NRGBD~\cite{azinovic2022neural}.}
  Models are evaluated with sparse sampling (one frame every two frames, -S). All recurrent rows use the same frozen CUT3R checkpoint; MeMix is applied to TTT3R (the setting of its paper), and TTSA3R is run by us.
  \colorbox{oursblue}{\strut Shaded rows} are our methods.
  \textbf{Bold} indicates the best result in each column at the same input length.
}
\label{tab:recon}
\setlength{\tabcolsep}{4pt}
\renewcommand{\arraystretch}{0.95}
\scriptsize
\begin{tabular}{lcccccccccccccc}
\toprule
& & \multicolumn{6}{c}{\textbf{7-Scenes-S}} & \multicolumn{6}{c}{\textbf{NRGBD-S}} \\
\cmidrule(lr){3-8}\cmidrule(lr){9-14}
& & \multicolumn{2}{c}{Acc $\downarrow$} & \multicolumn{2}{c}{Comp $\downarrow$} & \multicolumn{2}{c}{NC $\uparrow$}
  & \multicolumn{2}{c}{Acc $\downarrow$} & \multicolumn{2}{c}{Comp $\downarrow$} & \multicolumn{2}{c}{NC $\uparrow$} \\
\cmidrule(lr){3-4}\cmidrule(lr){5-6}\cmidrule(lr){7-8}
\cmidrule(lr){9-10}\cmidrule(lr){11-12}\cmidrule(lr){13-14}
\textbf{Model} & \textbf{$L$}
  & Mean & Med. & Mean & Med. & Mean & Med.
  & Mean & Med. & Mean & Med. & Mean & Med. \\
\midrule
VGGT~\cite{wang2025vggt} & 300/400/500 & \multicolumn{12}{c}{\textcolor{oomgray}{\textit{OOM at all tested lengths}}} \\
\midrule
\multirow{3}{*}{TTT3R~\cite{chen2025ttt3r}}
  & 300 & 0.039 & 0.023 & 0.023 & 0.005 & 0.565 & 0.600 & 0.104 & 0.045 & 0.026 & 0.005 & 0.605 & 0.668 \\
  & 400 & 0.050 & 0.029 & 0.026 & 0.005 & 0.557 & 0.587 & 0.144 & 0.066 & 0.068 & 0.009 & 0.592 & 0.643 \\
  & 500 & 0.065 & 0.037 & 0.031 & 0.006 & 0.550 & 0.575 & 0.166 & 0.087 & 0.087 & 0.016 & 0.588 & 0.637 \\
\midrule
\multirow{3}{*}{MeMix~\cite{dong2026memix}}
  & 300 & 0.034 & 0.020 & 0.023 & 0.005 & 0.567 & 0.603 & 0.099 & 0.037 & 0.020 & 0.004 & 0.616 & 0.692 \\
  & 400 & 0.043 & 0.025 & 0.026 & 0.005 & 0.560 & 0.590 & 0.146 & 0.066 & 0.070 & 0.018 & 0.602 & 0.665 \\
  & 500 & 0.059 & 0.032 & 0.030 & 0.005 & 0.553 & 0.580 & 0.183 & 0.094 & 0.094 & 0.031 & 0.595 & 0.650 \\
\midrule
\multirow{3}{*}{TTSA3R~\cite{zheng2026ttsa3r}}
  & 300 & 0.029 & 0.016 & 0.022 & 0.004 & \textbf{0.567} & \textbf{0.603} & 0.090 & 0.037 & 0.020 & 0.005 & 0.613 & 0.685 \\
  & 400 & 0.035 & 0.019 & 0.023 & 0.004 & 0.561 & 0.592 & 0.104 & 0.045 & 0.035 & 0.005 & 0.607 & 0.673 \\
  & 500 & 0.044 & 0.022 & 0.024 & 0.004 & 0.557 & 0.585 & 0.121 & 0.052 & 0.049 & 0.006 & 0.604 & 0.666 \\
\midrule
\multirow{3}{*}{AFG-Pose (Ours)}
  & \cellcolor{oursblue}300
    & \cellcolor{oursblue}0.025 & \cellcolor{oursblue}0.012
    & \cellcolor{oursblue}0.021 & \cellcolor{oursblue}\textbf{0.003}
    & \cellcolor{oursblue}0.566 & \cellcolor{oursblue}0.600
    & \cellcolor{oursblue}0.075 & \cellcolor{oursblue}0.026
    & \cellcolor{oursblue}\textbf{0.011} & \cellcolor{oursblue}0.004
    & \cellcolor{oursblue}\textbf{0.617} & \cellcolor{oursblue}0.690 \\
  & \cellcolor{oursblue}400
    & \cellcolor{oursblue}0.028 & \cellcolor{oursblue}0.014
    & \cellcolor{oursblue}0.022 & \cellcolor{oursblue}\textbf{0.003}
    & \cellcolor{oursblue}0.561 & \cellcolor{oursblue}0.592
    & \cellcolor{oursblue}0.085 & \cellcolor{oursblue}0.035
    & \cellcolor{oursblue}0.023 & \cellcolor{oursblue}\textbf{0.003}
    & \cellcolor{oursblue}0.613 & \cellcolor{oursblue}0.683 \\
  & \cellcolor{oursblue}500
    & \cellcolor{oursblue}0.031 & \cellcolor{oursblue}0.015
    & \cellcolor{oursblue}0.022 & \cellcolor{oursblue}\textbf{0.003}
    & \cellcolor{oursblue}0.558 & \cellcolor{oursblue}0.587
    & \cellcolor{oursblue}0.101 & \cellcolor{oursblue}0.040
    & \cellcolor{oursblue}0.033 & \cellcolor{oursblue}\textbf{0.004}
    & \cellcolor{oursblue}0.611 & \cellcolor{oursblue}0.680 \\
\midrule
\multirow{3}{*}{AFG-Img (Ours)}
  & \cellcolor{oursblue}300
    & \cellcolor{oursblue}\textbf{0.023} & \cellcolor{oursblue}\textbf{0.010}
    & \cellcolor{oursblue}\textbf{0.020} & \cellcolor{oursblue}\textbf{0.003}
    & \cellcolor{oursblue}0.563 & \cellcolor{oursblue}0.596
    & \cellcolor{oursblue}\textbf{0.071} & \cellcolor{oursblue}\textbf{0.023}
    & \cellcolor{oursblue}\textbf{0.010} & \cellcolor{oursblue}\textbf{0.003}
    & \cellcolor{oursblue}0.616 & \cellcolor{oursblue}\textbf{0.692} \\
  & \cellcolor{oursblue}400
    & \cellcolor{oursblue}\textbf{0.023} & \cellcolor{oursblue}\textbf{0.010}
    & \cellcolor{oursblue}\textbf{0.020} & \cellcolor{oursblue}\textbf{0.003}
    & \cellcolor{oursblue}0.558 & \cellcolor{oursblue}0.588
    & \cellcolor{oursblue}\textbf{0.075} & \cellcolor{oursblue}\textbf{0.026}
    & \cellcolor{oursblue}\textbf{0.015} & \cellcolor{oursblue}\textbf{0.003}
    & \cellcolor{oursblue}\textbf{0.617} & \cellcolor{oursblue}\textbf{0.689} \\
  & \cellcolor{oursblue}500
    & \cellcolor{oursblue}\textbf{0.024} & \cellcolor{oursblue}\textbf{0.010}
    & \cellcolor{oursblue}\textbf{0.021} & \cellcolor{oursblue}\textbf{0.003}
    & \cellcolor{oursblue}0.554 & \cellcolor{oursblue}0.581
    & \cellcolor{oursblue}\textbf{0.092} & \cellcolor{oursblue}\textbf{0.031}
    & \cellcolor{oursblue}\textbf{0.027} & \cellcolor{oursblue}\textbf{0.004}
    & \cellcolor{oursblue}\textbf{0.614} & \cellcolor{oursblue}\textbf{0.683} \\
\bottomrule
\end{tabular}
\end{table*}

\section{Qualitative 3D Reconstruction Comparison}
\label{app:recon_qual}

\begin{figure}[h]
\centering
\includegraphics[width=0.9\linewidth]{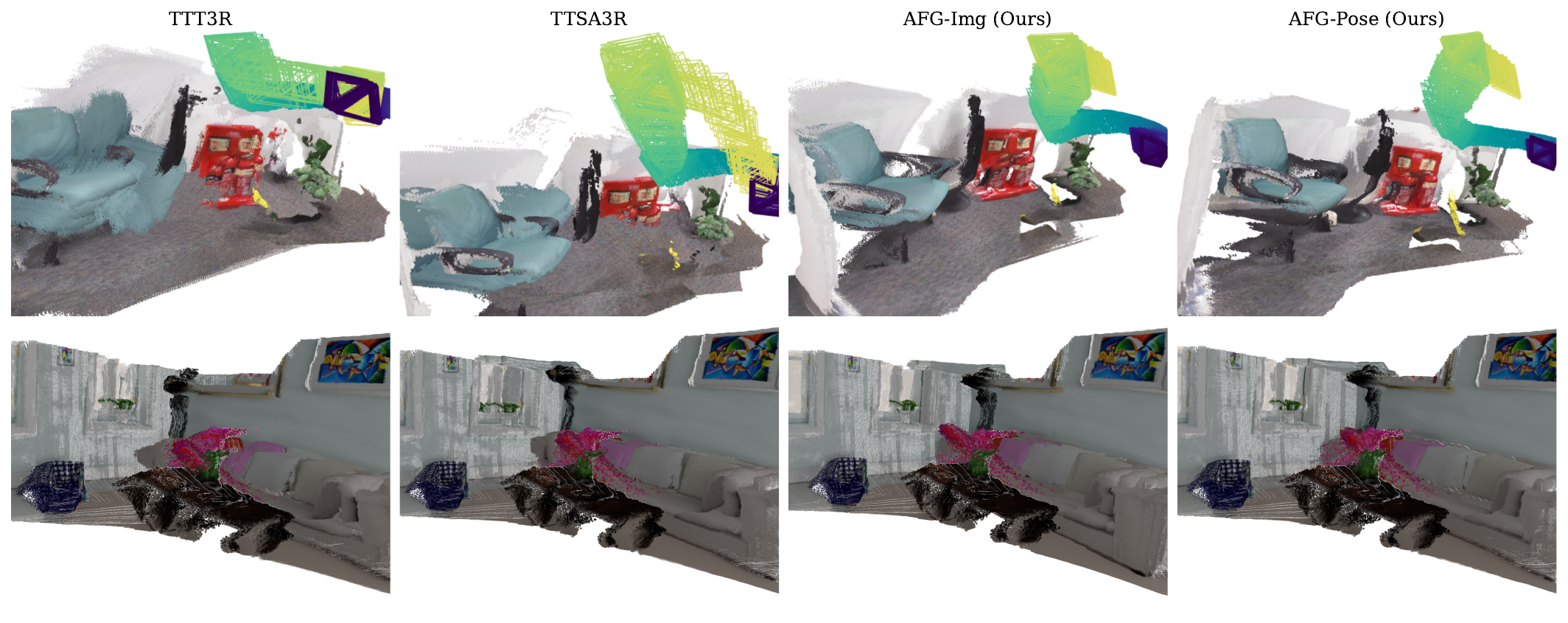}
\caption{\textbf{Qualitative 3D reconstruction comparison on 7-Scenes (top) and NRGBD (bottom).} Without adaptive gating, recurrent baselines accumulate drift and produce fragmented geometry. Top: TTT3R and TTSA3R smear the armchair and the red fire extinguisher, while both AFG variants keep them intact. Bottom: TTT3R shows duplicated, smeared geometry around the flowers and the sofa, which AFG reduces.}
\label{fig:vis_recon}
\end{figure}

Fig.~\ref{fig:vis_recon} provides qualitative point-cloud reconstructions on representative 7-Scenes~\cite{shotton2013scene} and NRGBD~\cite{azinovic2022neural} sequences, complementing the quantitative results in Table~\ref{tab:recon} (Appendix~\ref{app:recon_quant}). Without adaptive gating, recurrent baselines accumulate pose drift and produce fragmented geometry; AFG variants yield more coherent surfaces and better-preserved scene structure across both datasets.

\newpage

\end{document}